%% file: main.tex
\pdfoutput=1
\documentclass[11pt]{article}

\usepackage[final]{acl}

\usepackage{times}
\usepackage{latexsym}
\usepackage{amsmath,amsfonts,bm}
\usepackage{amssymb}% http://ctan.org/pkg/amssymb
\usepackage{pifont}% http://ctan.org/pkg/pifont

\usepackage[nodisplayskipstretch]{setspace}
\usepackage{booktabs}
\usepackage{multirow}
\usepackage[T1]{fontenc}
\usepackage[utf8]{inputenc}

\usepackage{microtype}

\usepackage{inconsolata}

\usepackage{graphicx}
\usepackage{hyperref}
\usepackage{url}
\usepackage{mathtools}
\usepackage[utf8]{inputenc}
\usepackage[T1]{fontenc}
\usepackage{booktabs} % for professional tables
\usepackage{amsfonts}
\usepackage{nicefrac}
\usepackage{microtype}
\usepackage{xcolor}
\usepackage{multirow}
\usepackage{subcaption}
\usepackage{xspace}
\usepackage{wrapfig}
\usepackage{colortbl}
\usepackage[most]{tcolorbox}
\usepackage{graphicx}
\usepackage{amsmath}
\usepackage[ruled, vlined]{algorithm2e}
\usepackage{qtree}

\input{macros}

\title{Injecting Syntactic Sub-Networks in Transformer Language Models with Tree Regularization \includegraphics[height=1em]{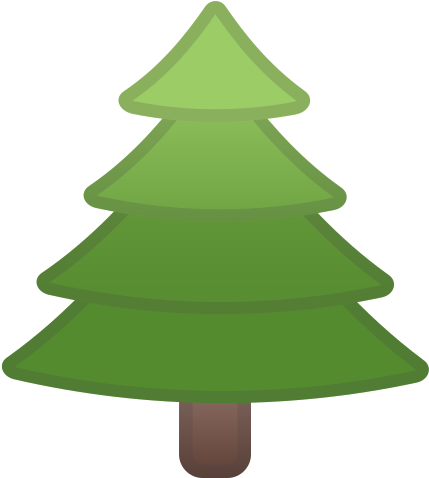}}
\title{Sneaking Syntax into Transformer Language Models \\ with Tree Regularization \includegraphics[height=1em]{emojipng.com-291948.png}}

\author{Ananjan Nandi, Christopher D. Manning \& Shikhar Murty\\
Department of Computer Science\\
Stanford University\\
Stanford, CA 94305, USA \\
\texttt{\{ananjan,smurty\}@stanford.edu} \\
}
\begin{document}
\maketitle
\begin{abstract}
While compositional accounts of human language understanding % in cognitive science 
are based on a hierarchical tree-like process, neural models like transformers lack a direct inductive bias for such tree structures. Introducing syntactic inductive biases could unlock more robust and data-efficient learning in transformer language models (LMs), but existing methods for incorporating such structure greatly restrict models, either limiting their expressivity or increasing inference complexity. This work instead aims to \emph{softly} inject syntactic inductive biases into given transformer \textit{circuits}, through a structured regularizer. We introduce \ours{}, an auxiliary loss function that converts bracketing decisions from silver parses into a set of differentiable orthogonality constraints on vector hidden states. $\ours{}$ integrates seamlessly with the standard LM objective, requiring no architectural changes. LMs pre-trained with $\ours{}$ on natural language corpora such as WikiText-103 achieve up to 10\% lower perplexities on out-of-distribution data and up to 9.5 point improvements in syntactic generalization, requiring less than half the training data to outperform standard LMs. \ours{} still provides gains for pre-trained LLMs: Continued pre-training of Sheared Llama with $\ours{}$ results in improved syntactic generalization, and fine-tuning on MultiNLI with \ours{} mitigates degradation of performance on adversarial NLI benchmarks by 41.2 points. We release all code to guide future research\footnote{\url{https://github.com/ananjan-nandi-9/tree_regularization}}.
 
%Transformer architectures lack a direct inductive bias for hierarchical structure. Therefore, in limited data settings, transformer language models (LMs), struggle with complex syntactic phenomena and compositionally novel sentences that humans can effortlessly understand. While prior work has improved syntactic generalization by incorporating structural inductive biases, this comes at the cost of additional parameters or inference complexity. In contrast, we introduce $\ours{}$, an auxiliary \emph{loss function} designed to softly bias transformers towards explicitly hierarchical structure by converting bracketing decisions in given constituency parses into orthogonality constraints on vector hidden states. $\ours{}$ integrates seamlessly with the LM objective, requiring no changes to the transformer architecture. LMs pre-trained with $\ours{}$ on natural language corpora such as WikiText-103 achieve up to 10\% lower perplexities on out-of-distribution data and up to 9.5 point improvements in syntactic generalization. Additionally, $\ours{}$ provides gains for models pre-trained at scale---continued pre-training of Sheared Llama with $\ours{}$ results in improved syntactic generalization, and fine-tuning on MultiNLI with $\ours{}$ mitigates degradation of performance by 41.2 points on adversarial NLI benchmarks.
\end{abstract}

\section{Introduction}

A substantial body of research \citep{crain1987structure, pallier2011cortical, van2013model, hale2018finding} suggests that human language processing is inherently \textit{hierarchical}: syntactic rules combine word-level meanings to give rise to semantics for larger constituents, which are then combined into sentences. In contrast, transformers \citep{vaswani2017attention}, the architecture behind large language models (LLMs), allow for arbitrary, non-hierarchical routing of information while processing an input sentence. 

LLMs are pre-trained on massive amounts of data to achieve reasonable generalization \cite{achiam2023gpt, dubey2024llama}, but recent studies find that state-of-the-art LLMs still struggle with compositional generalization \cite{guo2020hierarchical, berglund2023reversal, hale-stanojevic-2024-llms}, the ability to understand familiar words in novel contexts. Prior work \cite{socher2013recursive, eriguchi-etal-2016-tree} proposed explicitly tree-structured models capable of data-efficient compositional generalization, but they are too constrained and not amenable to pre-training at scale.

% [AN] Figure needs some changes? origin into the figure. Added.
\begin{figure*}[ht]
\setlength\belowcaptionskip{-5pt}
    \centering
    \includegraphics[width=\textwidth]{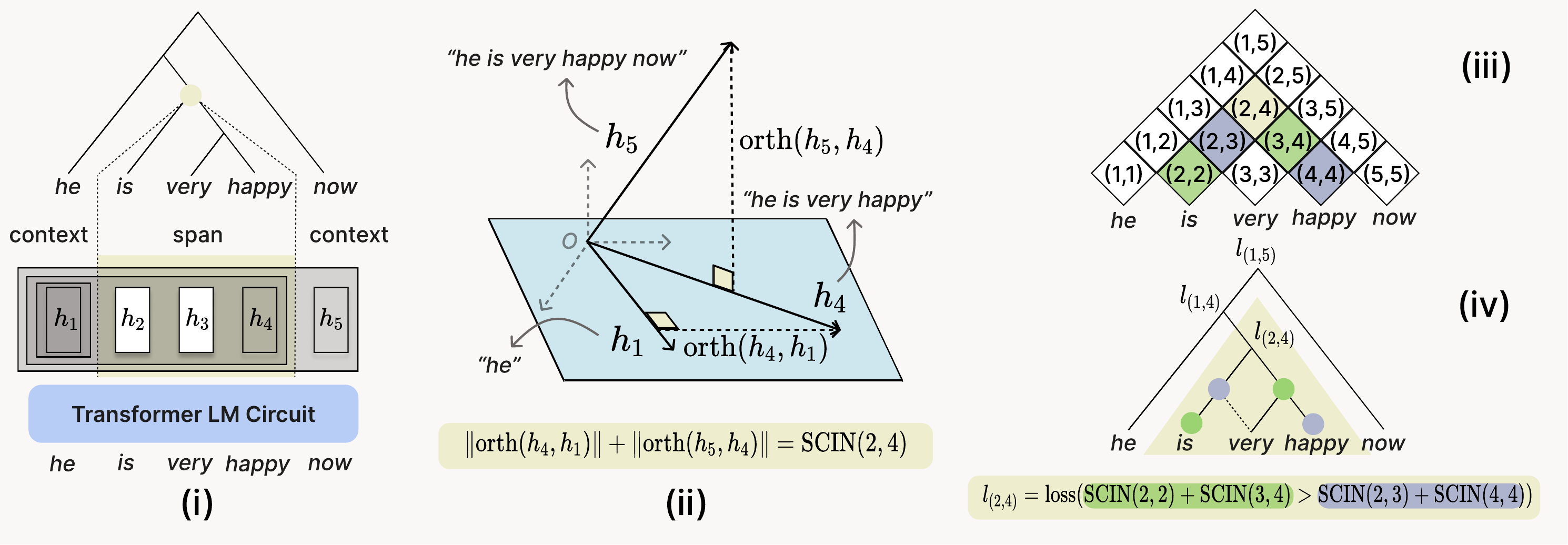}\vspace{-0.05in}
    \caption{$\ours{}$ loss ($\LMTR{}$) computation for $S = $ ``he is very happy now''. \textbf{(i)} Computation of vector hidden states $\cvec{h}_i$ by passing $S$ as input to some circuit of the LM. $\cvec{h}_i$ is the representation for the prefix of $S$ ending at $i$. \textbf{(ii)} Span Contextual Independence Score ($\SCIS{}$, \S~\ref{sec: SCI}) computation for ``is very happy''. Orthogonality constraints are enforced between span representation $\cvec{h}_4$ and its context $\cvec{h}_1$ and $\cvec{h}_5$. \textbf{(iii)} Chart of $\SCIS{}$ for all spans in $S$. \textbf{(iv)} Possible bracketings of ``is very happy'' are (``is very'', ``happy'') with score $\SCIS{}(2,3) + \SCIS{}(4,4)$ and (``is'', ``very happy'') with score $\SCIS{}(2,2) + \SCIS{}(3,4)$. Loss for this span ($l_{(2,4)}$) encourages the second bracketing. $\LMTR{} = l_{(1,5)} + l_{(1,4)} + l_{(2,4)}$ includes analogous losses for spans ``he is very happy'' and ``he is very happy now''.}
    \label{fig:desc}
\end{figure*}

Consequently, people have proposed inductive biases that encourage hierarchical computation in transformers, including Syntactic Language Models (SLMs, \citealp{sartran2022transformer, murty2023pushdown}) that model a joint distribution over words and syntax trees. While SLMs often improve data efficiency and syntactic generalization, they have several drawbacks: They often involve additional parameters to model syntax, constrain the attention mechanisms of the underlying model, or involve more complex and slower inference methodologies.

% . In parallel, \citet{murty2022characterizing} aims to investigate the extent to which the computation of a transformer is \emph{tree-structured} through \emph{representational invariance}: the contextual invariance of intermediate representations in a tree built over spans of the input. % how to better phrase this?

In this work, we instead devise a new differentiable loss function that \textit{softly} injects syntactic inductive biases into a given \textit{circuit} of the transformer: $\ours{}$. These biases are soft structural constraints that aim to ensure that the hidden state representation from the transformer circuit for each constituent in an input sentence is \emph{maximally orthogonal} to that of its surrounding context (Figure~\ref{fig:desc}). $\ours{}$ is simply added as a regularizer to the LM loss during training. Crucially, an LM trained with $\ours{}$ is \emph{completely indistinguishable} from a standard LM in both architecture and inference mechanism.

The use of $\ours{}$ improves syntactic generalization across model and data scales. On diagnostic sentence transformation tasks such as tense inflection and question formation \citep{mccoy2020does}, it significantly enhances grokking behavior (\S~\ref{sec:grokking}). It also improves syntactic generalization on the BLiMP \citep{warstadt2020blimp} and SyntaxGym \cite{gauthier2020syntaxgym} test suites by up to 3.2 and 9.5 points, both when used during pre-training from scratch (\S~\ref{sec:sent-level-lm}), and during continued pre-training (\S~\ref{sec:cpt}).  LMs with $\ours{}$ surpass the syntactic generalization of baseline LMs with less than half the training data (\S~\ref{sec:sample efficiency}).
Most importantly, \ours{} improves out-of-distribution language understanding. When used during pre-training of LMs on fully parsed (\S~\ref{sec:bllip}) as well as a mixture of parsed and unparsed (\S~\ref{sec:wikitext}) data, $\ours{}$ reduces out-of-distribution perplexities by up to 9.2\%.
When $\ours{}$ is used for fine-tuning an LLM on MultiNLI \citep{williams-etal-2018-broad}, it mitigates performance drops by up to 41.2 points on adversarial NLI benchmarks (\S~\ref{sec: NLI}). We release all code at \url{https://github.com/ananjan-nandi-9/tree_regularization}.

\section{Background and Notation}
\label{sec: bg}
\paragraph{Tree Structures in Language Processing.} Consider a sentence $S = \{x_1, x_2, \ldots, x_n\}$ where $x_j$ is the $j$\uprm{th} token in the sentence. Let tree $T(S)$ over sentence $S$ be a set of spans $\range{a}{b} = \{x_a, \ldots, x_b\}$ such that if $\range{a}{b}$ and $\range{b+1}{c} \in T(S)$, then $\range{a}{c} \in T(S)$. We say $\range{a}{c}$ is \textit{split} at index $b$ to get spans $\range{a}{b}$ and $\range{b+1}{c}$ of $T(S)$. Prior work \citep{pallier2011cortical, van2013model, hale2018finding} suggests that human language processing is hierarchical---sentences contain ``constituents'' or spans of words that act as semantic units, with smaller constituents (``is'', ``very happy'') recursively combining to form larger ones (``is very happy''), building up meaning incrementally. The constituency parse $\gparse{S}$ is the tree form by all constituents of $S$. Tree-structured models \citep{socher2013recursive, tai2015improved} build sentence representations through recursively bottom-up computation over constituency parses, whereas 
% transformers do not perform any explicit hierarchically-structured computation.
for transformers, any hierarchically structured computation is at most implicitly learned in the attention layers.

\paragraph{Hierarchical Structure in Transformers.} % How is hierarchical structure encoded in transformers. Talk about tree projections, and briefly say a few sentences about SLMs. 
Recent evidence suggests that pre-trained transformers behave similarly to tree-structured models on certain tasks \citep{murty2022characterizing, murty2023grokking}. The \textit{tree projection} metric from \citet{murty2022characterizing} quantifies how well a transformer's computation can be approximated by a tree-structured model. It is based on the idea that bottom-up computations on trees create context-invariant representations at intermediate nodes. Therefore, the tree built from maximally context-invariant spans aligns most closely with the transformer's computation. Another line of work builds Syntactic Language Models \cite{sartran2022transformer, murty2023pushdown, hu2024generative} that learn joint distributions over inputs and their parses to introduce syntactic inductive biases into transformer language models. These approaches involve substantial changes to the transformer architecture---explicitly modifying attention patterns, as well as changing the output space or inference mechanisms. $\ours{}$ provides an alternative, biasing transformer circuits towards hierarchical structure through regularization.

\paragraph{Evaluating Syntactic Generalization.} Syntactic generalization refers to an LM's ability to apply syntactic rules implicitly learned during training to new, unseen data. We use the BLiMP \citep{warstadt2020blimp} and SyntaxGym (SG, \citealp{gauthier2020syntaxgym}) test suites to evaluate LMs for syntactic generalization. In BLiMP, models are presented with pairs of grammatical and ungrammatical sentences that differ minimally, and are evaluated on assigning lower perplexity to the grammatical sentences. The SG test suites evaluate six distinct syntactic phenomena, with each suite involving a particular inequality constraint related to the surprisal of different continuations of input sentences based on their prefixes. These constraints are rooted in theories of syntax; for example, assigning higher surprisal to the third word in the sentence ``I know who you introduced to them.'' compared to ``I know that you introduced to them.''

\section{Our Approach}
% rework this
$\ours{}$ is a differentiable loss function that injects the hierarchical structure of input constituency parses into transformer circuits. Inspired by \citet{murty2022characterizing}, $\ours{}$ specifies a set of soft constraints to ensure that constituent representations are maximally independent of surrounding context. For any span, we quantify this independence with the Span Contextual Independence Score ($\SCIS{}$, \S~\ref{sec: SCI}). $\ours{}$ maximizes  $\SCIS{}$ for spans corresponding to input constituents while simultaneously minimizing it for other spans (\S~\ref{sec: loss}). In an optimally trained model, the tree with the highest cumulative $\SCIS{}$ thus recovers the constituency parse. We then provide a greedy algorithm to extract the parse tree induced by $\ours{}$ from a given model for any input sentence. A deeper discussion of the design choices underlying our method can be found in Appendix~\ref{sec:disc}.

% We can then probe for the parse trees used by the model by searching for the tree with the maximum cumulative $\SCIS{}$ using a greedy decoding algorithm. Additionally, we can leverage gold constituency parses to define a $\ours{}$ loss that increases $\SCIS{}$ for spans in the gold parse while reducing it for other spans, prompting the model to encode the gold parses in its intermediate representations.

\subsection{Span Contextual Independence Score}
\label{sec: SCI}
Let $\cvec{h}^{l}_{a,1}, \cvec{h}^{l}_{a,2}, \dots, \cvec{h}^{l}_{a,n}$ be $L_2$-normalized vector hidden states for each token in $S$ produced by attention head $a$ at layer $l$ of a causal auto-regressive transformer LM with multi-headed self-attention. Since most tasks benefit from a blend of hierarchical and unstructured computation, we only apply $\ours{}$ on the circuit formed by a given subset of attention heads $A$ at a given layer $l$.

\paragraph{Span vectors.} For any $j \leq n$, $\cvec{h}^{l}_{A,j}$ is the concatenation ($\#$) of vector representations from the attention heads in $A$ at layer $l$ for prefix $x_{\leq{j}}$:
\begin{align} 
  \cvec{h}^{l}_{A,j} = \#_{a \in A} \cvec{h}^{l}_{a,j}
\end{align}
Then, for span $\range{i}{j}$, $\cvec{h}^l_{A,i-1}$ contains information before the start of $\range{i}{j}$, or the \textit{context} for the span.

\paragraph{Contextual Independence.} We operationalize contextual independence in terms of orthogonality constraints between span representations. In particular, for $\range{i}{j}$ to be contextually invariant, we expect $\cvec{h}^{l}_{A,j}$ to be independent of, or \textit{orthogonal to} $\cvec{h}^{l}_{A,i-1}$. We also constrain $\cvec{h}^{l}_{A,j+1}$ to be orthogonal to $\cvec{h}^{l}_{A,j}$, as the information in $\range{i}{j}$ is expected to be independent of what comes after it in the sentence. Therefore, we define $\SCIS{}_{A}^{l}$ for $\range{i}{j}$ at layer $l$ for attention heads in $A$ as
\begin{align} 
  \mathrm{\SCIS{}}_{A}^{l}(i,j) &= \|\mathrm{orth}(\cvec{h}^{l}_{A,j}, \cvec{h}^{l}_{A,i-1})\| \nonumber\\ &+ \|\mathrm{orth}(\cvec{h}^{l}_{A,j+1}, \cvec{h}^{l}_{A,j})\|,
\end{align}

% \begin{equation} \label{eq:1}
% \begin{aligned}
%     SCIS(i,j) = \|\mathrm{orth}(h_j, h_{i-1})\| + \|\mathrm{orth}(h_{j+1}, h_j)\|
% \end{aligned}
% \end{equation}
\noindent
where $\|\|$ is the $L_2$ norm, and $\mathrm{orth}(x,y) = x - (x^{\top}y)y$ for any two normalized vectors $x$ and $y$. For the sake of completeness, $\|\mathrm{orth}(h_j, h_{-1})\|$  $\forall j$ and $\|\mathrm{orth}(h_{k+1}, h_k)\|$ are assumed to be 0. Since $l$ and $A$ are fixed during training, we simplify $\SCIS{}_{A}^{l}$ to $\SCIS{}$ from here on. 
% \begin{align} 
%   \mathrm{\SCIS{}}(T(S)) =  \sum_{\range{i}{j} \in T(S)} \mathrm{\SCIS{}}(i,j)
% \end{align}

% \begin{equation} \label{eq:2}
% \begin{aligned}
%     SCIS(T) =  \sum_{[i-j] \in T} SCIS(i,j)
% \end{aligned}
% \end{equation}

\subsection{\ours{} Loss}
\label{sec: loss}

Given sentence $S$, the $\ours{}$ loss ($\LMTR{}$) biases the model towards driving up $\SCIS{}$ for all spans in the constituency parse $\gparse{S}$ while driving down $\SCIS{}$ for other spans. Specifically, for constituent $\range{i}{j} \in \gparse{S}$ split at index $p$, we first compute split scores $\splitscore(i,q,j)$ for $i \leq q \leq j-1$ as
\begin{align} 
  \splitscore(i,q,j) &= \mathrm{\SCIS{}}(i,q) \nonumber \\ &+ \mathrm{\SCIS{}}(q+1,j).
\end{align}
We then use these scores to compute a span-level log loss $l_{(i,j)}$. These losses are computed as
\begin{align} 
  l_{(i,j)} = \log\bigg[{{\sum_{q=i}^{j-1} \exp{\splitscore(i, q,j)}}}\bigg] - \splitscore(i,p,j).  
  % \mathcal{L}_\text{CE}(\{\splitscore(i,q,j):q \in [i, j-1]\}, \nonumber \\ target = p)
\end{align}
$\LMTR{}$ is then the sum of span-level losses $l_{(i,j)}$ for all constituents in $\gparse{S}$,
\begin{align} 
  \LMTR{} =  \sum_{\range{i}{j} \in \gparse{S}} l_{(i,j)}.
\end{align}

In practice, $\LMTR{}$ is computed recursively top-down on $\gparse{S}$ following Alg.~\ref{alg:greedyloss}. $\LMTR{}$ is added as an auxiliary loss to the LM loss during training, resulting in the training objective $\LMLoss{} + \alpha \LMTR{}$ where $\LMLoss{}$ is the LM loss. Since both $\LMTR{}$ and $\LMLoss{}$ are cross-entropy losses, $\alpha$ can generally be set to 1. $\LMTR{}$ and $\LMLoss{}$ can also consume data from different datasets. For example, we can perform LM on a large pre-training dataset while passing batches from a smaller, parsed dataset to $\LMTR{}$. 

\paragraph{Recovering parses during inference.}
\label{sec:probe} During inference, we can use a top-down greedy decoding algorithm to recover the unique parse tree encoded in the $\ours{}$ constraints for any given $S$ from the hidden states of a circuit. Given $\SCIS{}$ scores for all spans in $S$, we recover the tree that maximizes the sum of $\SCIS{}$ scores across its spans.  Specifically, the split for any span $\range{i}{j}$ happens at the index $\hat{p}$ that maximizes $\splitscore(i,q,j)$,
\begin{align} 
  \hat{p} = \argmax_{q \in [i,j-1]}{\splitscore(i,q,j)}.
\end{align}
$\range{i}{\hat{p}}$ and $\range{\hat{p}}{j}$ can then be recursed on to obtain more constituents of this \emph{induced} parse tree. Details of this method can be found in Alg.~\ref{alg:greedy}.

\subsection{Implementation Details}
\label{sec:impl}

% [AN] while TreeReg theoretically results in the addition of another quadratic loss term during training, it is efficient in practice, due to the following factors ...

$\SCIS{}$ computation does not require additional model calls and can be performed during the forward pass that computes $\LMLoss{}$. $\SCIS{}$ scores for all spans in a sentence can be calculated simultaneously and efficiently using vectorization, as detailed in Appendix~\ref{sec:vectorization}. Additionally, updates from $\LMLoss{}$ and $\LMTR{}$ can occur at different frequencies during training, for example, $\LMTR{}$ may be applied once every $k$ steps of $\LMLoss{}$.
Despite the quadratic time complexity of computing $\LMTR{}$ (which scales with the square of input token count), these factors collectively ensure that $\ours{}$ remains efficient in practice. As per our implementation, there is an increase of around $25\%$ in training time if $\LMTR{}$ is applied once every 10  $\LMLoss{}$ steps, on $25\%$ of attention heads. This excludes the time required for parsing the data used in $\ours{}$. The parses are either included with the dataset (BLLIP-LG, MultiNLI) or can be computed once and reused for every training run.

\section{Warm-up: Improving Grokking on Sentence Transformation Tasks}
\label{sec:grokking}

We start by training transformer LMs on two diagnostic tasks derived from PCFGs (See Appendix~\ref{sec:datasetstats} for examples and data statistics). In Tense Inflection (TI), the model is provided with an input in the past tense, and required to generate the same input in the present tense. In Question Formation (QF), the model is required to transform a declarative sentence into a question. For QF, we report first word accuracy of the decoded question, and for TI, we report the fraction of test inputs for which the target verb is correctly inflected.

 \paragraph{Setup.} We train 4-layer transformer LMs for 500k steps (Base LM), performing extended training to enable grokking as reported in prior work \citep{murty2023grokking}. For $\ours{}$ LM, we auto-parse both datasets with the Berkeley Neural Parser (Benepar, \citealp{kitaev2022learned}), and use $\LMTR{}$ on 2 out of 8 attention heads using hidden states from layer 2 of the LM, once every 20 steps of $\LMLoss{}$. To account for variance across training runs, we report averages as well as best performance at the end of training across 4 seeds (Table~\ref{tab: grokking}). We also report the training iteration (averaged over all runs) after which the performance on the test set converges.

\paragraph{Results.} LMs trained with $\ours{}$ grok faster and achieve higher performance than Base LMs on average, across datasets. Notably, on QF, $\ours{}$ LM obtains a 57.5 pt gain over a standard LM (generalizing perfectly) while also grokking over 10 times faster on average. 

\begin{table}[h!]
\setlength\belowcaptionskip{-5pt}
\centering
\small
\renewcommand{\arraystretch}{1.2}
\resizebox{\columnwidth}{!}{%
\begin{tabular}{@{}lccc@{}}
\toprule
\textbf{Model} & \textbf{Avg. Acc. ($\uparrow$)} & \textbf{Best Acc. ($\uparrow$)} & \textbf{itr. ($\downarrow$)} \\ \midrule
\hrow \multicolumn{4}{c}{\it Tense Inflection (TI)} \\
Base LM & 47.2 $\pm$ 16.7 & 71.1 & 427k $\pm$ 41k \\
$\ours{}$ LM & \textbf{90.4 $\pm$ 6.3} & \textbf{98.3} & \textbf{391k $\pm$ 35k} \\
\hrow \multicolumn{4}{c}{\it Question Formation (QF)} \\
Base LM & 42.1 $\pm$ 15.4 & 66.9 & 460k $\pm$ 7k \\
$\ours{}$ LM & \textbf{99.6 $\pm$ 0.7} & \textbf{100.0} & \textbf{43k $\pm$ 26k} \\
\bottomrule
\end{tabular}
}
\caption{Evaluations on TI and QF tasks. We report averaged test accuracy across seeds (\textbf{Avg. Acc.}), the best test accuracy (\textbf{Best Acc.}), and the average iteration after which test performance converges (\textbf{itr.}). $\ours{}$ LMs grok faster and achieve better performance. }
\label{tab: grokking}
\end{table}

\begin{table*}[h!]
\setlength\belowcaptionskip{-5pt}
\centering
\small
\renewcommand{\arraystretch}{1.2}
\resizebox{\textwidth}{!}{%
\begin{tabular}{@{}lcccccc@{}}
\toprule
\multirow{2}{*}{\textbf{Model}} & \multicolumn{2}{c}{\textbf{Syntactic Generalization ($\uparrow$)}} & \multicolumn{2}{c}{\textbf{PPL ($\downarrow$)}} & \multirow{2}{*}{\textbf{Modified Architecture}} & \multirow{2}{*}{\textbf{Inference Overhead}}\\ \cmidrule(l){2-3}\cmidrule(l){4-5} 
                                & \textbf{BLiMP} & \textbf{SG} & \textbf{BLLIP} & \textbf{PTB} &  &  \\ \midrule
Base LM & 72.2 & 71.9 & 21.6 & 49.1 & - & -  \\ \midrule
\textsc{PLM} & 75.1 & 80.2 & 29.8 & - & \xmark & \cmark \\ 
\textsc{TG} & - & \textbf{82.5} & 30.3 & - & \cmark & \cmark \\
\textsc{Pushdown LM} & \textbf{75.6} & 82.3 & \textbf{19.9} & - & \cmark & \cmark \\ \midrule
$\ours{}$ LM & \textbf{74.8} & \textbf{80.0} & 22.3 & \textbf{44.6} & \xmark & \xmark \\ \bottomrule
\end{tabular}%
}
\caption{Syntactic Generalization on BLiMP and SG
test suites, and perplexities (\textbf{PPL}) on BLLIP-LG test set (\textbf{BLLIP}) and PTB test set (\textbf{PTB}) for models trained on BLLIP-LG. Results for \textsc{PLM}, \textsc{TG} and \textsc{Pushdown LM} are taken from~\citealp{murty2023pushdown}. $\ours{}$ LMs show better syntactic generalization and better generalization to PTB compared to the Base LM.}
\label{tab:baselines}
\end{table*}

\section{Sentence-Level Language Modeling}\label{sec:sent-level-lm}

Can $\ours{}$ provide gains when introduced in pre-training? We investigate this under settings where data passed to $\LMLoss{}$ and $\LMTR{}$ is the same (\S~\ref{sec:bllip}) and different (\S~\ref{sec:wikitext}).

\subsection{Language Modeling on BLLIP-LG}
\label{sec:bllip}

\paragraph{Setup.} We train 16-layer LMs on the BLLIP-LG dataset \citep{hu-etal-2020-systematic} (Base LM, detailed hyperparameters in Appendix~\ref{sec:hyperparams}). For $\ours{}$ LM, we use $\LMTR{}$ at layer 12 of the LM, on 2 out of 8 attention heads, once every 10 steps of $\LMLoss{}$, on the already-parsed BLLIP-LG dataset. A discussion of the layer and attention heads used for $\LMTR{}$ can be found in Appendix~\ref{sec:ablation}. We benchmark syntactic generalization using the BLiMP and SG test suites (See \S~\ref{sec: bg}). To evaluate out-of-distribution generalization, we report perplexity on the Penn TreeBank (PTB; \citealp{marcus1993building}) test set.

To the best of our knowledge, all prior approaches to induce hierarchical computations in autoregressive causally-trained transformer language models do not leave the underlying transformer architecture untouched, and thus are not directly comparable with \ours{}.
Nevertheless, we also report results for Parsing as Language Model (\textsc{PLM};~\citealp{qian2021structural}), Transformer Grammars (\textsc{TG};~\citealp{sartran2022transformer}), and Pushdown Layers (\textsc{Pushdown LM};~\citealp{murty2023pushdown}) in Table~\ref{tab:baselines}.

\begin{figure}[h]
\setlength\belowcaptionskip{-5pt}
    \centering
    \includegraphics[width=\columnwidth]{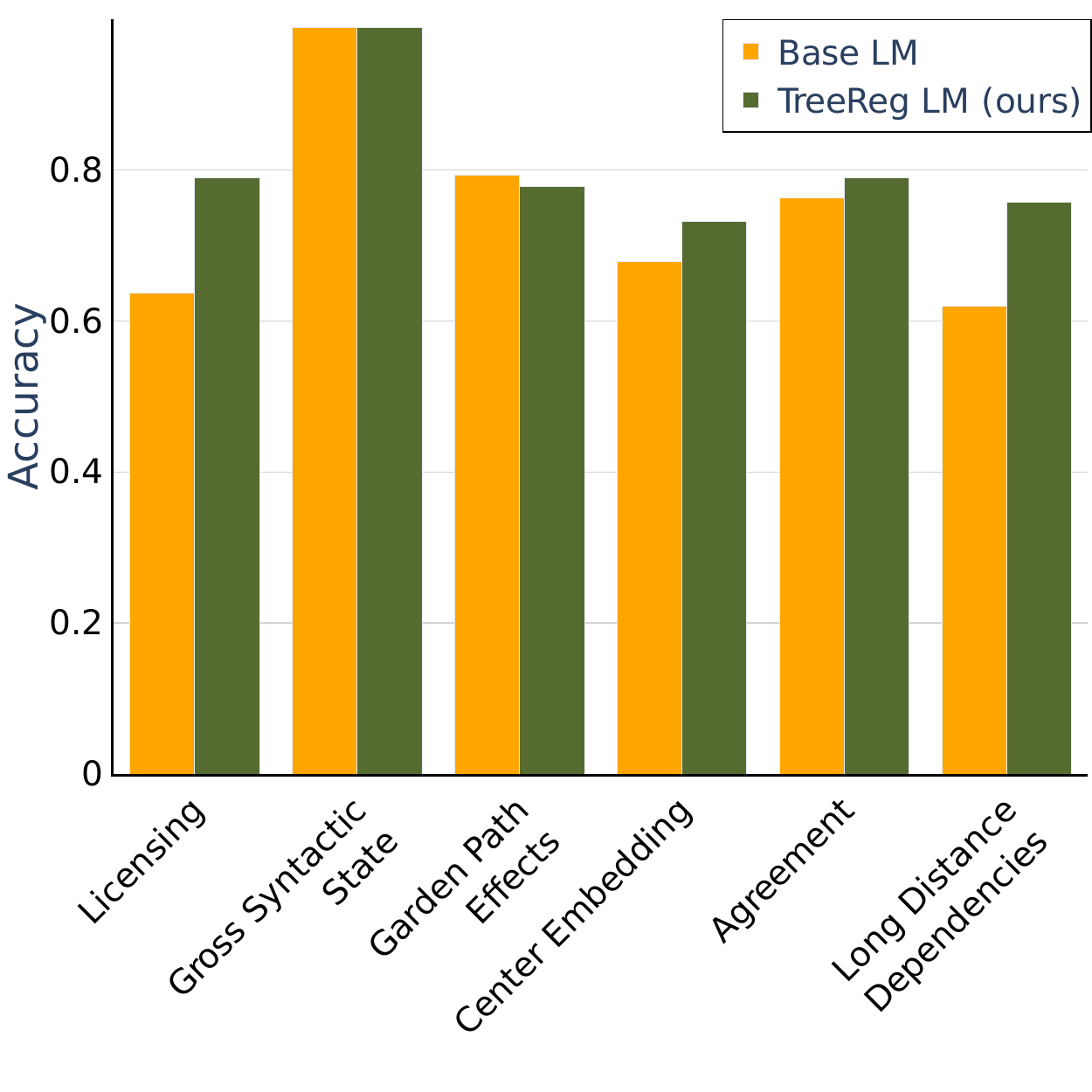}\vspace{-0.05in}
    \captionsetup{skip=0pt}
    \caption{Comparing $\ours{}$ LM with Base LM from Table~\ref{tab:baselines} on SG test suites. $\ours{}$ LM outperforms the Base LM on 4 out of 6 test suites, with 1 tie.}
    \label{fig:sg} 
\end{figure}

\paragraph{Results.} $\ours{}$ achieves a 2.6 point (pt) gain on BLiMP and 8.1 pt overall gain on SG test suites. Figure~\ref{fig:sg} shows $\ours{}$ improves over the baseline on 4 out of 6 SG test suites, with a notable 17 pt gain on Licensing. 
% Appendix~\ref{} contains a more in-depth discussion of results on the SG test suites. % might have to remove/send to appendix.
$\ours{}$ LM also generalizes better to PTB with 9.2\% lower perplexity, with a marginal 0.7 pt perplexity increase on BLLIP-LG.

% Add average lines

While \textsc{PLM}, \textsc{TG} and \textsc{Pushdown LM} outperform \ours{} at syntactic generalization on BLiMP and SG test suites, they require additional parameters or incur inference overheads. Except for \texttt{Pushdown LM}, these models also have substantially higher perplexities on the BLLIP-LG test set compared to \ours{}, showing the benefits of the unchanged transformer architecture.

\subsection{Sample Efficiency}
\label{sec:sample efficiency}
\begin{figure}[h!]
    \setlength\belowcaptionskip{-5pt}
    \centering
    \includegraphics[width=0.48\textwidth]{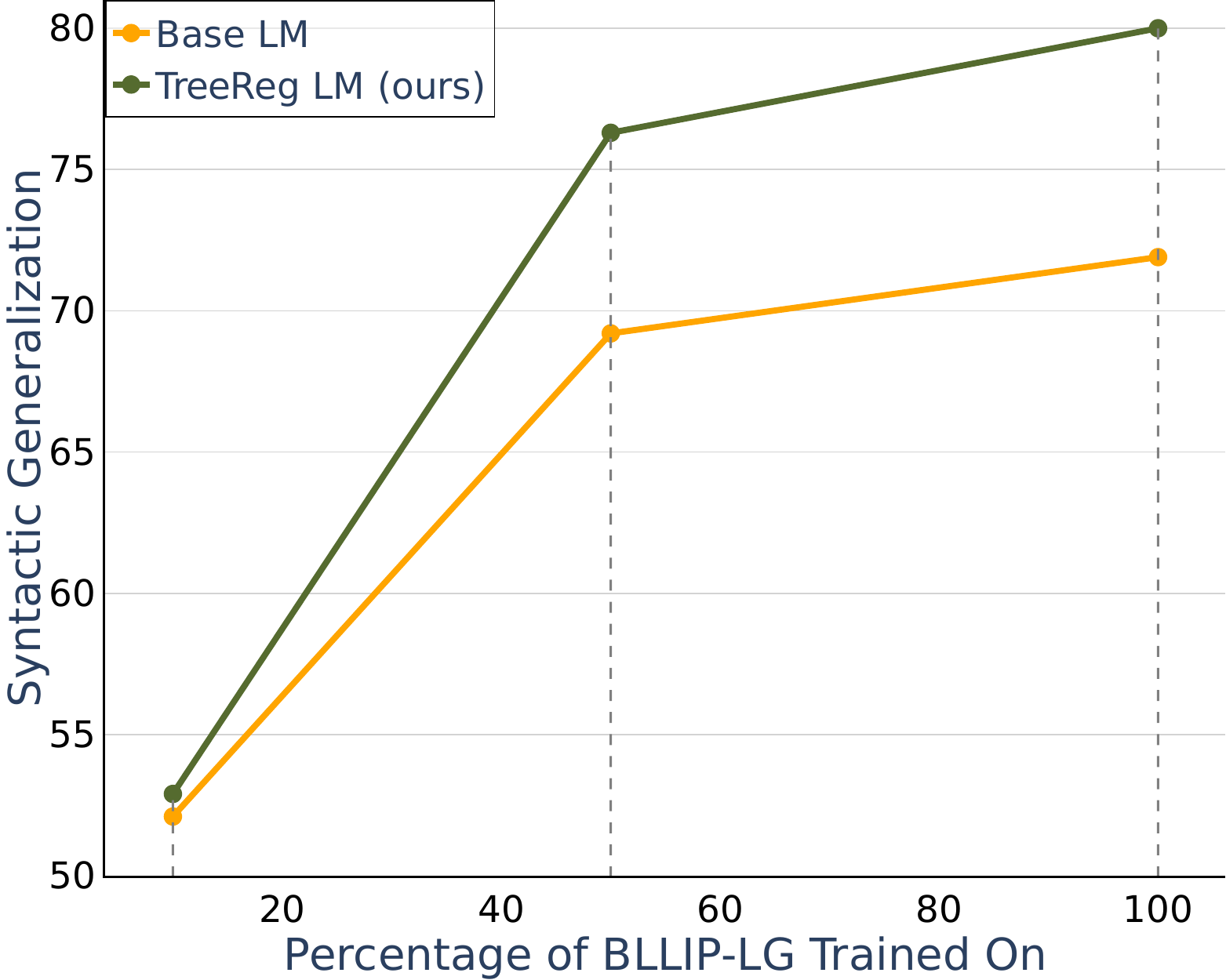}\vspace{-0.05in}
    \captionsetup{skip=10pt}
    \caption{Plot of Syntactic Generalization on SG test suites vs Percentage of BLLIP-LG data used to train LMs from scratch. $\ours{}$ LM exceeds the maximum syntactic generalization performance of Base LM with less than 50\% of the data.}
    \label{fig:sgse}
\end{figure}

\paragraph{Setup.} To measure sample efficiency gains from $\ours{}$, we train 16-layer LMs with the same configuration as \S~\ref{sec:bllip} on [10, 50, 100]\% of BLLIP-LG. Only parses of the data used for LM is used for $\ours{}$ in each case. We present syntactic generalization results on SG in Figure~\ref{fig:sgse}.

\paragraph{Results.} While syntactic generalization improves with more training data for both the Base and $\ours{}$ LMs, $\ours{}$ LMs outperform the baseline across all settings. Moreover, the performance gap widens as the amount of training data increases. Notably, $\ours{}$ surpasses the baseline's maximum syntactic generalization performance  on SG by 4.4 points when trained on only 50\% of the data, demonstrating substantially more sample-efficient syntactic generalization. Therefore, even though $\ours{}$ increases training time by 25\% compared to a baseline trained for the same number of iterations (see \S~\ref{sec:impl}), it reaches better syntactic generalization in only 62.5\% of the baseline's training time.

\subsection{Language Modeling on WikiText}
\label{sec:wikitext}

\paragraph{Setup.} Auto-parsing text corpora may not always be practical for large-scale language modeling, particularly when the corpus is very large or when the text lacks clear syntactic structure, as is the case for web data. Therefore, we experiment with a setting where the pre-training and parsed data come from different data sources. We train a 12-layer GPT-2-small model (Base LM) on chunks of 1024 tokens from WikiText-103 \cite{merity2022pointer} with the exact hyperparameters and tokenization of GPT-2 small (with the addition of dropout ($=0.1$) to prevent overfitting). For $\ours{}$ LM, we reuse the BLLIP-LG parses and use $\LMTR{}$ at layer 4, on 3 out of 12 attention heads, once every 10 steps of $\LMLoss{}$. We also test a setting where LM training is additionally performed on one batch from BLLIP-LG for every 20 batches from WikiText-103 (Table~\ref{tab: wikitext}) for both Base LM and $\ours{}$ LM\@. PTB is again used to evaluate out-of-distribution generalization.

% Pushdown LM tables
% Plotly figures. Times New Roman.

\begin{table}[h]
\setlength\belowcaptionskip{-5pt}
\centering
\small
\renewcommand{\arraystretch}{1.2}
\resizebox{\columnwidth}{!}{%
\begin{tabular}{@{}lcccc@{}}
\toprule

\multirow{2}{*}{\textbf{Model}} & \multicolumn{2}{c}{\textbf{Syntactic Generalization ($\uparrow$)}} & \multicolumn{2}{c}{\textbf{PPL ($\downarrow$)}} \\ \cmidrule(l){2-3}\cmidrule(l){4-5} 
                                & \textbf{BLiMP} & \textbf{SG} & \textbf{WikiText} & \textbf{PTB} \\ \midrule
\hrow \multicolumn{5}{c}{\it No LM on BLLIP-LG} \\
Base LM & 71.0 & 69.4 & \textbf{17.5} & \textbf{331.5} \\
$\ours{}$ LM & \textbf{72.5} & \textbf{73.7} & 17.8 & 411.1 \\ 
\hrow \multicolumn{5}{c}{\it LM on BLLIP-LG} \\
Base LM & 71.5 & 67.2 & \textbf{18.3} & 53.2 \\
$\ours{}$ LM & \textbf{74.1} & \textbf{76.7} & 19.7 & \textbf{50.8} \\ \bottomrule
\end{tabular}%
}
\caption{Syntactic Generalization on BLiMP and SG
test suites, and perplexities on the WikiText and PTB test sets for GPT-2-small models trained on WikiText, with and without LM on BLLIP-LG. $\ours{}$ LMs show better syntactic generalization in both settings.}
\label{tab: wikitext}
\end{table}

\paragraph{Results.} We observe improvements in syntactic generalization with $\ours{}$ in both settings, obtaining up to 2.6 pt gain on BLiMP and 9.5 pt overall gain on SG test suites. The gains are relatively smaller when LM training is not performed on BLLIP-LG, and perplexities on PTB increase by 24\% over the baseline. In this setting, we find that perplexities are also high on BLLIP-LG, averaging around 300. As a result, we conjecture that the hidden states used in $\ours{}$ are not adequately trained in this setting, leading to worse out-of-distribution perplexities.

In the other setting, Base LM struggles to utilize the LM training on BLLIP-LG, resulting in a 2.2 pt overall drop on SG. In contrast, $\ours{}$ shows an overall increase of 3 pt on SG test suites. Perplexities on PTB are again 2.4 pts lower with $\ours{}$ in this setting compared to Base LM, showing better generalization from the BLLIP-LG LM training.

\section{Continued Pre-training and Fine-tuning}

Next, we apply $\ours{}$ to LLMs. In particular, we use the Sheared Llama-1.3B model of \citet{xiasheared} and explore $\ours{}$ under continued pre-training and fine-tuning settings.

\subsection{Continued Pre-training on BLLIP-LG}
\label{sec:cpt}
\paragraph{Setup.} We perform continued pre-training of Sheared Llama-1.3B (Base LM) on BLLIP-LG, with a context window of 512 tokens, to get CPT LM. For $\ours{}$ LM, BLLIP-LG parses are used, and $\LMTR{}$ is applied once every 5 steps of $\LMLoss$, on 6 out of 24 attention heads at layer 16 of the LLM. We report scores on the BLiMP and SG test suites and perplexity on PTB (Table~\ref{tab:bllipcpt}).

\begin{table}[h]
\setlength\belowcaptionskip{-5pt}
\centering
\small
\renewcommand{\arraystretch}{1.2}
\resizebox{\columnwidth}{!}{%
\begin{tabular}{@{}lcccc@{}}
\toprule

\multirow{2}{*}{\textbf{Model}} & \multicolumn{2}{c}{\textbf{Syntactic Generalization ($\uparrow$)}} & \multicolumn{2}{c}{\textbf{PPL ($\downarrow$)}} \\ \cmidrule(l){2-3}\cmidrule(l){4-5} 
                                & \textbf{BLiMP} & \textbf{SG} & \textbf{BLLIP} & \textbf{PTB} \\ \midrule
Base LM & 73.6 & 80.7 & 23.8 & 42.5 \\ \midrule
CPT LM & 80.7 & 83.9 & \textbf{9.24} & 15.2 \\
$\ours{}$ CPT LM & \textbf{81.9} & \textbf{85.5} & 9.41 & \textbf{14.3} \\ \bottomrule
\end{tabular}%
}
\caption{Syntactic Generalization on BLiMP and SG
test suites, and perplexities on BLLIP-LG and PTB test sets for Sheared Llama-1.3B (\textbf{LM}) continued pre-trained (\textbf{CPT}) on BLLIP-LG. With $\ours{}$, CPT LM shows better syntactic generalization and PTB perplexities.}
\label{tab:bllipcpt}
\end{table}

\paragraph{Results.} $\ours{}$ results in a 1.2 pt gain on BLiMP and 1.6 pt overall gain on SG test suites, along with a 0.9 pt decrease in PTB perplexity. These improvements are less significant than those reported in \S~\ref{sec:bllip}, which we conjecture is due to the greater challenge in rewiring the unstructured inductive biases learned by the LLM during large-scale pre-training.

\subsection{Fine-tuning on MultiNLI}
\label{sec: NLI}

Can $\ours{}$ prevent LLMs fine-tuned on classification tasks from learning spurious shortcuts? To answer this, we consider Natural Language Inference (NLI), a task that involves classifying the logical relationship between a \textit{premise} and a \textit{hypothesis} sentence as \textit{entailment}, \textit{contradiction}, or \textit{neutral}. Prior work \citep{mccoy2019right, geiger2020neural} suggests that models often learn spurious shortcuts when trained on broad coverage NLI datasets, causing them to fail on diagnostic out-of-distribution datasets.

\paragraph{Setup.} We finetune Sheared Llama-1.3B (Base LM) on MultiNLI \cite{williams-etal-2018-broad} (training details and evaluation methodology in Appendix~\ref{sec:nliapp}) to get FT LM. For $\ours{}$ LM, we binarize the parses of the premise and hypothesis sentences provided in the dataset and apply $\LMTR{}$ at layer 16 on 6 out of 24 attention heads, every 10 steps of $\LMLoss{}$. We report accuracy on the MultiNLI test set, which includes a \textit{matched} split from the same sources as the training data, and a \textit{mismatched} split from different sources. We also present results (Table~\ref{tab: nli}) on two diagnostic datasets: MoNLI \cite{geiger2020neural} and MED \cite{yanaka2019can}.

\begin{table}[h]
\setlength\belowcaptionskip{-5pt}
\centering
\small
\renewcommand{\arraystretch}{1.2}
\resizebox{\columnwidth}{!}{%
\begin{tabular}{@{}lcccc@{}}
\toprule

\multirow{2}{*}{\textbf{Model}} & \multicolumn{2}{c}{\textbf{MultiNLI ($\uparrow$)}} & \multicolumn{2}{c}{\textbf{Adversarial ($\uparrow$)}} \\ \cmidrule(l){2-3}\cmidrule(l){4-5} 
                                & \textbf{Matched} & \textbf{Mismatched} & \textbf{MoNLI} & \textbf{MED}\\ \midrule
Base LM & 35.5 & 35.1 & 50.3 & 50 \\ \midrule
FT LM & 65.9 & 66.3 & 1.9 & 43.1 \\
$\ours{}$ FT LM & \textbf{68.1}  & \textbf{68.0} & \textbf{43.5} & \textbf{45.8} \\ \bottomrule
\end{tabular}%
}
\caption{NLI accuracies on MultiNLI test splits and two \textbf{Adversarial} NLI evaluation datasets for Sheared Llama-1.3B (\textbf{LM}) finetuned (\textbf{FT}) on MultiNLI. $\ours{}$ results in more gains on MultiNLI and lower decreases on the adversarial benchmarks.}
\label{tab: nli}
\end{table}

\paragraph{Results.} We note improvements of up to 2.2 pt on MultiNLI from \ours{}. Next, we find a decrease of 48.4 pt on MoNLI and 6.9 pt on MED when Base LM is finetuned on MultiNLI. With $\ours{}$, the finetuned model is substantially more robust, with a moderate decrease of 6.8 pt on MoNLI and 4.2 pt on MED compared to Base LM. We conclude that  $\ours{}$ discourages the LM from learning spurious shortcuts for this task.

\section{Analysis}
\label{sec:analysis}

\paragraph{Parsing.} $\ours{}$ induces a tree-structured inductive bias in transformer LMs.
We explore how well these induced trees align with given constituency parses. In particular, we use the procedure detailed in \S~\ref{sec:probe} to recover induced parses from $\ours{}$ LM (\S~\ref{sec:bllip}) and $\ours{}$ CPT LM  (\S~\ref{sec:cpt}) trained on BLLIP-LG. We evaluate parsing on the auto-parsed BLLIP-LG and the manually annotated PTB test sets. Additionally, to estimate parsing on domains other than Newswire, we evaluate parsing on the 4000 Questions dataset \cite{judge2006questionbank}. Since $\ours{}$ induces unlabeled binarized parses, we report unlabeled F1 scores against binarized parses in Table~\ref{tab: parsing}. 

% maybe not enough trees to fully introduce structure into Sheared Llama weights. Rerun with WikiTrees

% Parse vs length graphs in appendix
% Parsing errors in appendix

\begin{table}[h]
\setlength\belowcaptionskip{-5pt}
\centering
\small
\renewcommand{\arraystretch}{1.2}
\resizebox{\columnwidth}{!}{%
\begin{tabular}{@{}lccc@{}}
\toprule
\textbf{Model} & \textbf{BLLIP ($\uparrow$)} & \textbf{PTB ($\uparrow$)} & \textbf{4kQ ($\uparrow$)} \\ \midrule
$\ours{}$ LM & \textbf{95.2}  & 88.4 & 91.6 \\
$\ours{}$ CPT LM & 94.6 & 89.1 & \textbf{94.5} \\
\midrule
\citealp{kitaev2022learned} & - & \textbf{94.7} & 87.7 \\
\bottomrule
\end{tabular}
}
\caption{Unlabeled F1 scores against parses from the BLLIP-LG, PTB and 4000 Questions (4kQ) test sets for $\ours{}$ LMs trained from scratch (\textbf{$\ours{}$ LM}) and continued pre-trained from Sheared Llama-1.3B (\textbf{$\ours{}$ CPT LM}) on BLLIP-LG. F1 scores are near or above 90 for all datasets. We also present unlabeled F1 scores from Benepar \cite{kitaev2022learned} for comparison.}
\label{tab: parsing}
\end{table}

Although $\ours{}$ only implicitly encodes constituency parses in transformer hidden states, we observe that it demonstrates remarkable parsing capabilities. Parses induced by $\ours{}$ align closely with given parses on all datasets, with F1 scores near or above 90. Parsing performance on PTB and 4000 Questions is slightly better with $\ours{}$ CPT LM, likely due to improved  representations from Sheared Llama-1.3B (which benefits from large-scale pre-training) on these datasets, that are relatively out-of-distribution from the BLLIP-LG training data. Notably, induced parses from \ours{} align better with silver parses than  Benepar~\citep{kitaev2022learned} on the 4000 Questions datasets, by 6.8 F1 pts. We provide some example parses on each dataset in Appendix~\ref{sec:parses}.

\begin{figure}[h]
\setlength\belowcaptionskip{-5pt}
    \centering
    \includegraphics[width=0.5\textwidth]{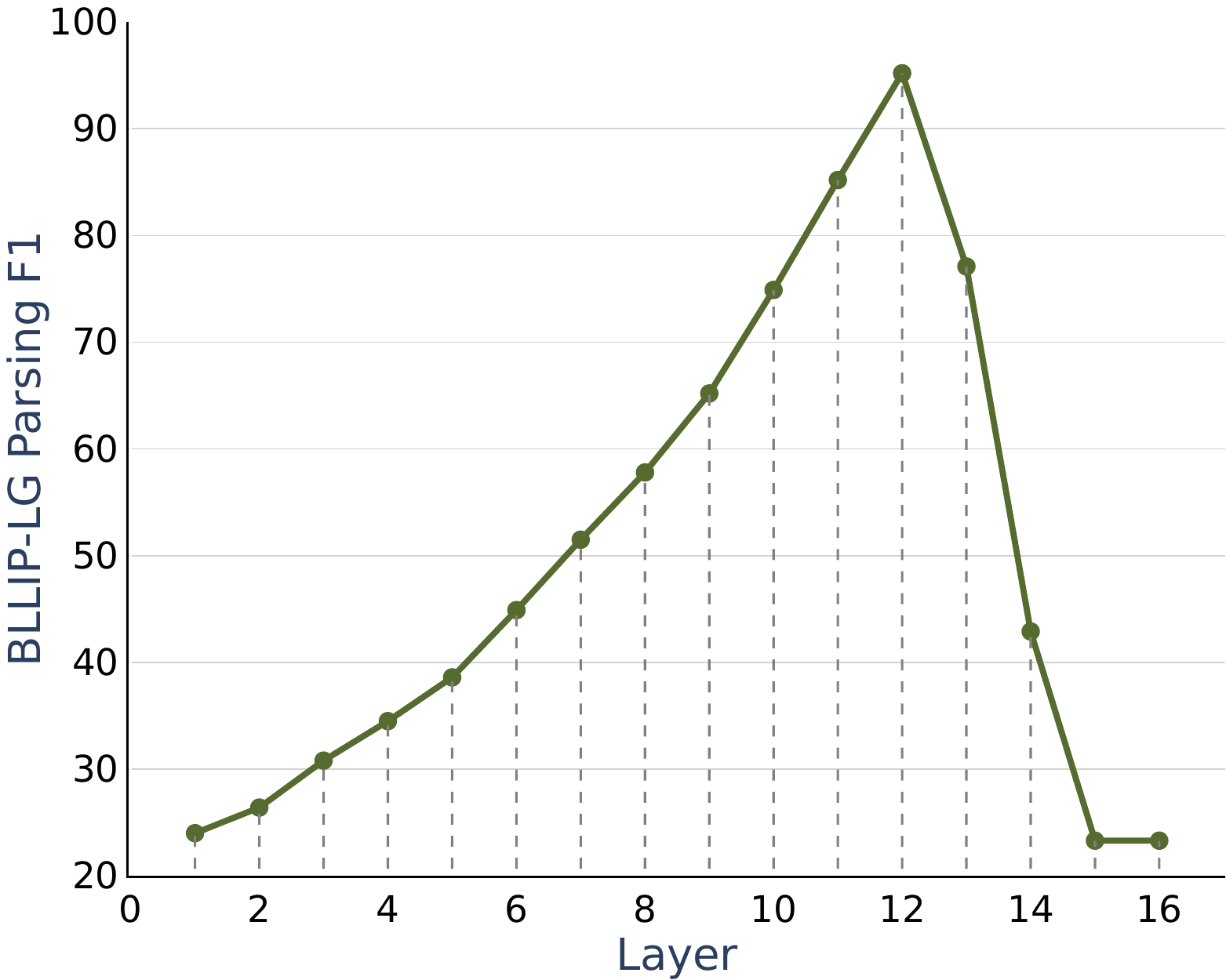}\vspace{-0.05in}
    \captionsetup{skip=10pt}
    \caption{Unlabeled F1 scores on the BLLIP-LG test set for parse trees induced from every layer of a 16-layer $\ours{}$ LM trained on BLLIP-LG with $\LMTR{}$ at layer 12. Circuits become increasingly tree-structured till layer 12, then rapidly become unstructured.}
    \label{fig:parse}
\end{figure}

\paragraph{Emergent structure across layers.} We investigate the evolution of tree structure across layers in models trained with $\ours{}$. To do this, we first infer trees for the BLLIP-LG test set from the circuit formed by the attention heads used to train $\ours{}$, at each layer of the 16-layer $\ours{}$ LM from \S~\ref{sec:bllip} (trained with $\LMTR{}$ at layer 12). We report unlabeled F1 scores of parses recovered at every layer against given parses in Figure~\ref{fig:parse}. 

% look at examples that were parsed correctly vs incorrectly, run mixed-effects regression for SG, evaluate parses using Benepar, see if parses are consistent first

The F1 scores reveal that at layer 1, the parses are nearly random. As we progress to layer 12, they increasingly align with given parses. Beyond layer 12, there is a sharp drop in F1 scores back to near-random by layer 16. This suggests that the circuit involved in $\LMTR{}$ becomes increasingly aligned with given parses up to layer 12, and then rapidly loses this structure. We conclude that to perform next-word prediction, $\ours{}$ LMs implement a non-syntactic function over the syntactic representations built at layer 12. These syntactic representations are, in turn, built gradually over the preceding layers of the transformer, even in the absence of explicit supervision from $\ours{}$ at these layers. 
% In Appendix~\ref{sec:rand}, we also show that improvements from $\ours{}$ come specifically from hierarchical computation that align closely with given parses.

\paragraph{Training on Randomized Parses.}
To assess if the gains from $\ours{}$ are specifically due to hierarchical computation that align closely with constituency parses, we conduct a controlled experiment. We train a 16-layer LM using the exact setup described in \S~\ref{sec:bllip}, but instead of using gold parses, we provide $\ours{}$ with random parses of BLLIP-LG sentences (Randomized $\ours{}$ LM). These randomized parses are generated top-down recursively, by choosing random split points for each sentence span. We then compare performance with the $\ours{}$ LM and Base LM on the BLiMP and SG test suites (Table~\ref{tab: rand}).

\begin{table}[h]
\setlength\belowcaptionskip{-5pt}
\centering
\small
\renewcommand{\arraystretch}{1.2}
\resizebox{\columnwidth}{!}{%
\begin{tabular}{@{}lcc@{}}
\toprule
\textbf{Model} & \textbf{BLiMP ($\uparrow$)} & \textbf{SG ($\uparrow$)} \\ \midrule
Base LM & 72.2  & 71.9 \\
$\ours{}$ LM & \textbf{74.8} & \textbf{80} \\
Randomized $\ours{}$ LM & 73.4 & 71.8 \\
\bottomrule
\end{tabular}
}
\caption{Results on BLiMP and SG test suites for LMs trained from scratch on BLLIP-LG, with silver and randomized parses (\textbf{Randomized $\ours{}$ LM}) passed to $\ours{}$. Performance decreases with randomized parses compared to silver parses.}
\label{tab: rand}
\end{table}

As expected, overall performance on the SG test suites drops marginally compared to Base LM when randomized parses are fed to $\ours{}$. On the other hand, BLiMP accuracies increase by 1.2 pts even in this setting.  We hypothesize that this discrepancy arises from 
BLiMP's less granular sentence-level perplexity-based evaluation, in contrast to the surprisal-based approach used in SG.

\begin{figure}[h]
\setlength\belowcaptionskip{-5pt}
    \centering
    \includegraphics[width=0.5\textwidth]{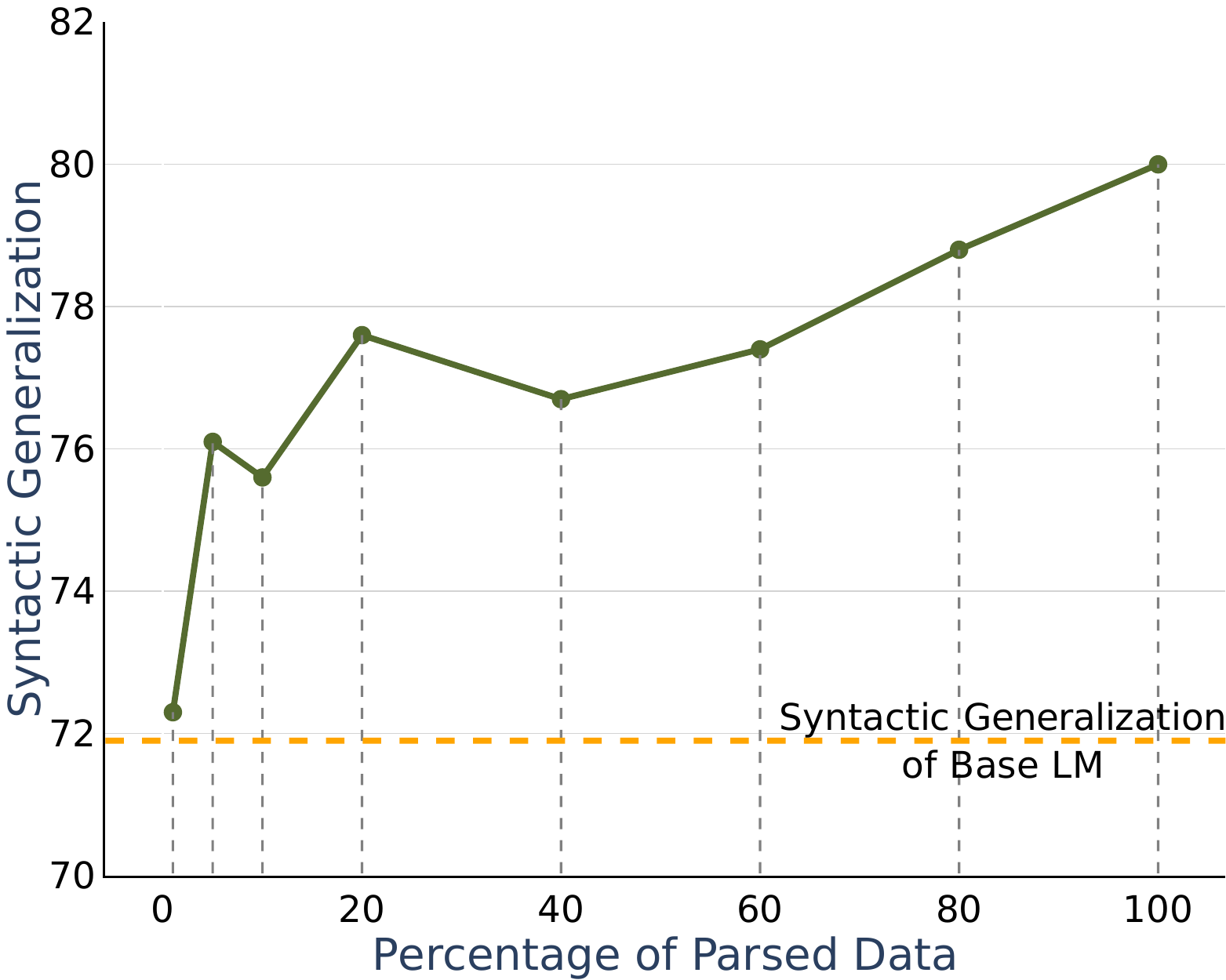}\vspace{-0.05in}
    \captionsetup{skip=10pt}
    \caption{Syntactic Generalization on SG test suites vs Percentage of parsed BLLIP-LG data provided to $\ours{}$, for LMs trained from scratch on BLLIP-LG. Even with 1\% of the data, $\ours{}$ LMs have better syntactic generalization than baseline LMs.}
    \label{fig:sgtest}
\end{figure}

\paragraph{Dependence on amount of parsed data.} We train 16-layer LMs on BLLIP-LG using $\ours{}$, with the same setup as \S~\ref{sec:bllip}, but vary the percentage of parsed BLLIP-LG used for $\ours{}$ across [1, 5, 10, 20, 40, 60, 80, 100]\%. We report the overall SG test suite performance for these settings in Figure~\ref{fig:sgtest}.

% \begin{figure*}[h]
%     \centering
%     \includegraphics[width=0.5\textwidth]{ptb_prop.png}\vspace{-0.05in}
%     \caption{PennTreeBank Parsevals vs \% of BLLIP-LG passed to $\ours{}$ (Trained from Scratch)}
%     \label{fig:sg}
% \end{figure*}

As the amount of parsed data provided to $\ours{}$ increases, we see a general trend of improvement in syntactic generalization on the SG test suites. Remarkably, we see improvements over Base LM from \ours{} even when it is provided with parses for only 1\% of the data used for LM.

%Remarkably, with merely 1\% of the data parsed, $\ours{}$ show superior syntactic generalization compared to the baseline.

% \begin{figure*}[h]
%     \centering
%     \includegraphics[width=0.5\textwidth]{blimp_train_prop.png}\vspace{-0.05in}
%     \caption{BLiMP Accuracies vs \% of BLLIP-LG used for LM + $\ours{}$ training (Trained from Scratch)}
%     \label{fig:sg}
% \end{figure*}

% BLLIP-LG for now, replace with GPT-2 on WikiText

\section{Other Related Work}
\label{sec: related work}

% Mech Interpretability

\paragraph{Generalization Failures of Transformer LMs.} A substantial body of research has pointed out the limitations of transformer-based LMs in achieving robust compositional generalization. They have been shown to struggle with simple tasks, such as determining the parity of binary strings \citep{bhattamishra-etal-2020-ability, chiang-cholak-2022-overcoming}, balancing bracket sequences \citep{hahn2020theoretical}, as well as more complex diagnostic natural language tasks \citep{mccoy2020does, geiger2020neural, NEURIPS2023_deb3c281}. Large-scale pre-training does not fully resolve these issues. For instance, \citet{berglund2023reversal} shows that GPT-4 \citep{achiam2023gpt} fails to generalize on bijective relationships (for example, answering ``Who is Tom Cruise's mother?'' correctly but not ``Who is Mary Lee South's son?'').~\citealp{hale-stanojevic-2024-llms} also find that Gemini Pro~\citep{team2023gemini} does not learn syntactic universals such as the Final-over-Final condition~\citep{sheehan2017final} for low-resource languages such as Basque.

\paragraph{Tree-structure in Neural Networks.} Building on theories of tree-structured human language processing, prior work explores various tree-structured model architectures \citep{socher2013recursive, tai2015improved, le-zuidema-2015-compositional, dyer-etal-2016-recurrent, shen2018ordered, hu-etal-2021-r2d2}. These models often excel at data-efficient compositional generalization due to their inherently hierarchical computation. However, they have been largely overshadowed by non-hierarchical pre-trained transformer-based LLMs \citep{brown2020language, achiam2023gpt}.

\paragraph{Linguistic Structure in Transformers.} One promising alternative to tree-structured models is to inject syntactic inductive biases into transformers either by jointly modeling sentences and parse structure \citep[among others]{qian2021structural, sartran2022transformer, murty2023pushdown}, or via constraints on self-attention \citep{strubell-etal-2018-linguistically, wang-etal-2019-tree, deshpande-narasimhan-2020-guiding, sartran2022transformer}. Although these models bring data efficiency and generalization improvements, they often require additional parameters, impose rigid constraints on attention mechanisms, or complicate inference. To the best of our knowledge, $\ours{}$ is the first approach to introduce explicit, albeit soft, syntactic biases into the transformer without modifying the architecture. This is achievable because standard transformers can encode hierarchical languages of bounded depth \citep{yao-etal-2021-self} when trained appropriately.

\section{Conclusion}
We propose $\ours{}$, a structured regularizer that injects soft syntactic inductive biases into given circuits of a transformer LM, converting constituency parses of input sentences into differentiable orthogonality constraints on vector hidden states. Without requiring any architectural changes to the transformer, models pre-trained with $\ours{}$ achieve up to 10\% lower perplexities on out-of-distribution data and enhance syntactic generalization by up to 9.5 points on standard test suites. When applied to pre-trained LLMs, $\ours{}$ improves syntactic generalization during continued pre-training and mitigates degradation on adversarial NLI benchmarks by up to 41.2 points when used during fine-tuning. $\ours{}$ more than doubles the sample efficiency of syntactic generalization and does not require the entire training dataset to be parsed, making it practical and efficient. To the best of our knowledge, this work is among the first to translate insights from a mechanistic interpretability-related work \citep{murty2022characterizing} into an actionable modeling innovation. We leave unsupervised alternatives to $\ours{}$, pre-training at scale with $\ours{}$, and application of $\ours{}$ to languages other than English as future work.

\section*{Limitations}
$\ours{}$ requires constituency-parsed datasets, which might be difficult to obtain for languages other English due to the lack of easily available constituency parsers. This also makes $\ours{}$ inapplicable to languages that do not have constituency structure, and we limit the experiments presented here to English. $\ours{}$ also introduces some additional hyperparameters that require tuning, such as the layer and subset of attention heads on which $\ours{}$ is be applied. As a heuristic, applying $\ours{}$ at the middle layer of the model, on 25\% of the attention heads, tends to work well across settings (see our ablations in Appendix~\ref{sec:ablation} for more details). Finally, while we optimize $\ours{}$ computation through vectorization, it still adds some computational overhead during training.

\bibliography{custom}

% Acknowledgements
% Chris group and NLP group, Derek, Martijn, Harshit

\appendix

\section{Discussion of Design Decisions}
\label{sec:disc}

\subsection{Last Hidden State as Representation for Prefix}
\label{sec:hiddenstates}
% [AN] present evidence that last hidden states might be doing some sort of summarization, easily computable

We use the hidden state for the last token in a span as an easily computable representation of the span. There is some evidence suggesting that in the middle layers of an autoregressive transformer language model, the burden of computation of representations tends towards the right, indicating that intermediate hidden states summarize corresponding prefixes to some extent. For instance,~\citealp{allen2023physics} find that when a transformer is trained on generations from a Probabilistic Context-Free Grammar (PCFG), ancestors of non-terminals are encoded at the hidden state of the last token in the span produced from the non-terminal. Therefore, we leverage the prefix information summarized in the hidden state of the last token of a span and rely on the LM's substantial expressive capabilities to embed $\ours{}$'s orthogonality constraints into its hidden states.

\subsection{Orthogonality for Operationalizing Contextual Independence}
\label{sec:orthogonality}
% [AN] emphasize that our way of operationalizing contextual independence is fast and efficient, compared to other ways such as Tree Projections, and that is a soft constraint that can be overriden to some extent by the LM loss
% [AN] Discuss why norm(orth): cosine similarity does not work as well
% [AN] Mention that the objective might make it seem like all non-constituent representations should be degenerate (norm(orth) = 0). However, this is not the case in practice, and the LM loss is the reason why.

We choose to operationalize contextual independence between spans as orthogonality of correspoding hidden states, to design an efficiently computable loss function.
There exist other ways of operationalizing contextual independence but these are inefficient and often require multiple forward passes on the transformer. For instance,~\citealp{murty2022characterizing} defines contextual independence as the similarity between the sum of hidden states for a span with unconstrained attention, compared to those from a transformer with masked attention for the prefix of the span. Another efficient alternative is using the negative absolute value of cosine similarity between span vector representations to represent contextual independence. This does not work as well, resulting in a syntactic generalization of $76.4$ on SG test suites (a drop of $3.6$ points) and $70.6$ on BLiMP (a drop of $4.2$ points) when used during the pretraining of a 16-layer LM from scratch on BLLIP-LG (\S~\ref{sec:bllip}).

TreeReg softly encourages higher $\SCIS{}$ for constituents of the parse tree, and lower $\SCIS{}$ for other constituents. However, a low $\SCIS{}$ score indicates greater similarity between span vectors, due to a smaller orthogonal component between them, which can result in degenerate hidden states. In practice, this is mitigated by opposing pressure from $\LMLoss{},$ that encourages divergent hidden states for different tokens in the sentence. To confirm this, we compute the average $\SCIS{}$ for constituent and non-constituent spans in the PennTreeBank test dataset, for the Base LM and \ours{} LM trained in \S~\ref{sec:bllip}. For the \ours{} LM, the average $\SCIS{}$ for constituents is $1.29$, and that for non-constituents is $0.98$. In contrast, for the Base LM, the average $\SCIS{}$ for constituents is $0.66$, and that for non-constituents is $0.65$.
We therefore conclude that the average $\SCIS{}$ for non-constituents is substantially positive for a model trained with \ours{}, and in fact is even higher than that for a model trained without \ours{}.

\subsection{Computation of $\SCIS{}$}
\label{sec:futuremasking}
% [AN] Discuss the future-masking constraint orth(j+1,j). Self-containment of information, this is soft as well.

Our method for computation of $\SCIS{}$ involves the addition of two terms, the first one being $\|\mathrm{orth}(\cvec{h}_{j}, \cvec{h}_{i-1})\|$, which captures the contextual independence between the information in a span and its prefix. The second term is $\|\mathrm{orth}(\cvec{h}_{j+1}, \cvec{h}_{j})\|$, which is a measure of how much the suffix of a span is dependent on the information contained in the span. The information in constituent spans should be mostly self-contained in a tree-structured computation, and therefore the hidden state after the span is expected to be independent of it. While this could result in enforced orthogonalities between consecutive words that may belong to the same constituent, \ours{} only enforces soft constraints, and there is opposing pressure from \LMLoss{} mitigating any undesirable behavior.
We also found that \ours{} improves syntactic generalization even without this future-masking constraint, reaching $77.4$ on SG test suites (a drop of $2.6$ points) and $72.8$ on BLiMP (a drop of $2.0$ points) when used during pre-training of a 16-layer LM from scratch on BLLIP-LG (\S~\ref{sec:bllip}).

\subsection{Greedy Decoding for Recovering Parses}
\label{sec:greedyparsing}
% [AN] Mention that TreeReg does not supervise on spans that cross constituents. Global parsing algorithms are not likely to work well since the SCIN of such spans are functionally untrained.

An alternative to the proposed top-down greedy decoding algorithm for recovering induced parse trees (\S\ref{sec:probe}) is a dynamic programming approach like the CKY algorithm~\citep{cocke1969programming, kasami1966efficient, younger1967recognition}. However, while computing $\LMTR{}$ during training, we only apply supervision on spans that are possible splits of constituents in the silver parse tree. For example, when applying $\ours{}$ to “the quick brown fox jumped over the river”, there is no supervision on the $\SCIS{}$ for the span “fox jumped over”, as “fox” is a part of the constituent “the quick brown fox”, and “jumped over” is a part of the constituent “jumped over the river”. A globally optimal parsing algorithm such as CKY will also consider $\SCIS{}$ for spans that had no supervision on similar spans during training, and these can be arbitrarily large, resulting in worse parsing performance.

\section{Algorithms for Computing $\ours{}$ Loss and Induced Parse Trees}
\label{sec:algs}

Alg.~\ref{alg:greedyloss} details the procedure used to compute $\ours{}$ loss for a given sentence. Alg.~\ref{alg:greedy} details the procedure used to recover induced parsed trees from $\ours{}$-trained models.

\begin{algorithm}[h!]
\setlength\belowcaptionskip{-5pt}
\SetNoFillComment
\caption{$\ours{}$ Loss}
\label{alg:greedyloss}
    \SetKwInOut{Input}{Input}
    \SetKwInOut{Output}{Output}

    function loss$(SCIN,P,i,j)$\;
    \tcp{Start: loss$(SCIN,P,0,n)$}
    \Input{
    
    $SCIN$ : Dictionary of $\SCIS{}$ for all spans

            $P$: Dictionary of split points of constituents in $\gparse{S}$
    
            $i$: Starting index of span $\range{i}{j}$
            
            $j$: Ending index of span $\range{i}{j}$}
    \Output{$l_{i,j}$, contribution to $\LMTR{}$ from $\range{i}{j}$}

    \If{$j - i < 1$}  
      {
        \tcc{If span has length 1 or 2, only one possible split}
        return $0$;
      }
      
    scores = $[SCIN(i,q) + SCIN(q+1,j)$ for $q \in [i,j-1]]$

    p = $P(\range{i}{j})$

    span\_loss = $cross\_entropy(scores, p)$

    left\_loss = loss$(SCIN,P,i,p)$
    
    right\_loss = loss$(SCIN,P,p+1,j)$

    \Return left\_loss + span\_loss + right\_loss
\end{algorithm}

\begin{algorithm}[h!]
\setlength\belowcaptionskip{-5pt}
\SetNoFillComment
\caption{$\ours{}$ Probe}
    \label{alg:greedy}
    \SetKwInOut{Input}{Input}
    \SetKwInOut{Output}{Output}

    \texttt{function} parse$(SCIN,i,j)$\;
    \tcp{Start: parse$(SCIN,0,n)$}
    \Input{
    
    $SCIN$ : Dictionary of $\SCIS{}$ for all spans,
    
            $i$: Starting index of span $\range{i}{j}$
            
            $j$: Ending index of span $\range{i}{j}$}
    \Output{Constituents in parse tree induced by $\SCIS{}$ for $\range{i}{j}$}

    \If{$i == j$}  
      {
        \tcc{If span has length 1, terminate recursion}
        return $\range{st}{st}$;
      }

    split\_point = $argmax_{q \in [i,j-1]}$ $SCIN(i,q) + SCIN(q+1,j)$

    left\_constituents = parse$(SCIN,i,split\_point)$
    
    right\_constituents = parse$(SCIN,split\_point + 1, j)$

    \Return $\range{i}{j}$ + left\_constituents + right\_constituents
\end{algorithm}

\section{Efficient Computation of $\SCIS{}$}
\label{sec:vectorization}

In this section, we detail our approach to efficiently compute $\SCIS{}$ for all spans in a sentence simultaneously through vectorization. For sentence $S$, let $\textbf{H}$ be the tensor containing all hidden states $\cvec{h}^{l}_{A,j}$:
\begin{align} 
  \textbf{H} = [\cvec{h}^{l}_{A,1}, \ldots, \cvec{h}^{l}_{A,n}]
\end{align}
Note that $\textbf{H}$ has dimension $(n,kd)$, where $d$ is the dimensionality of the hidden states from a single attention head of the model. We create a tensor $\textbf{O}$ whose elements correspond to orthogonals dropped from vectors in $\textbf{H}$ to every other vector in $\textbf{H}$, with dimensionality $(n,n,kd)$. In PyTorch, this can be implemented as follows:
\begin{align} 
  \textbf{H'} = \textbf{H}\mathrm{.unsqueeze(dim=1)} \\
  \textbf{D} = \mathrm{torch.sum(}\textbf{H} * \textbf{H'}\mathrm{, dim=-1)} \\
  \textbf{O} = \textbf{H} - \textbf{D}\mathrm{.unsqueeze(dim=-1)} \mathrm{*} \textbf{H'}
\end{align}
We then have:
\begin{align} 
  \mathrm{orth}(\cvec{h}^{l}_{A,j}, \cvec{h}^{l}_{A,i}) = \textbf{O}[i,j,:]
\end{align}
and:
\begin{align} 
  \SCIS{}(i,j) = \|\textbf{O}[i-1,j,:]\| \nonumber \\
  + \|\textbf{O}[j,j+1,:]\|
\end{align}
where $\|\|$ is the $L_2$ norm as before.

\section{Dataset Statistics}
\label{sec:datasetstats}

In this section, we provide statistics for all datasets used in our experiments in Table~\ref{tab: stats}. We also provide some examples from the Tense Inflection, Question Formation, MultiNLI, MoNLI and MED datasets.

\begin{table}[ht]
\setlength\belowcaptionskip{-5pt}
\centering
\small
\renewcommand{\arraystretch}{1.2}
\resizebox{\columnwidth}{!}{%
\begin{tabular}{@{}lrrr@{}}
\toprule
\textbf{Dataset} & \textbf{Train} & \textbf{Val} & \textbf{Test} \\ \midrule
Tense Inflection &  100,000$^{\dag}$  &  1,000$^{\dag}$ & 10,000$^{\dag}$ \\
Question Formation &  100,000$^{\dag}$  &  1,000$^{\dag}$ & 10,000$^{\dag}$ \\
BLLIP-LG & 42,000,000$^*$  & 36,000$^*$ & 72,000$^*$ \\
WikiText-103 & 1,801,350$^{\dag}$ & 3,760$^{\dag}$ & 4,358$^{\dag}$ \\
MultiNLI & 392,702$^{\dag}$ & 20,000$^{\dag}$ & 20,000$^{\dag}$ \\
\bottomrule
\end{tabular}
}
\caption{Statistics for each dataset used in our paper. {$^\dag$}: Number of data points. *: Number of tokens when tokenized using the GPT-2 tokenizer.}
\label{tab: stats}
\end{table}

\subsection{Tense Inflection Examples}
\begin{itemize}
    \setlength\itemsep{0em}
    \item \textbf{Input.} my quail upon my peacocks \textbf{remembered} her unicorns.
    \item \textbf{Output.} my quail upon my peacocks \textbf{remembers} her unicorns.
\end{itemize}

\begin{itemize}
    \setlength\itemsep{0em}
    \item \textbf{Input.} the vultures upon my unicorn \textbf{waited}. 
    \item \textbf{Output.} the vultures upon my unicorn \textbf{wait}.
\end{itemize}

\subsection{Question Formation Examples}
\begin{itemize}
    \setlength\itemsep{0em}
    \item \textbf{Input.} my raven that doesn't sleep \textbf{does} change.
    \item \textbf{Output.} \textbf{does} my raven that doesn't sleep change?
\end{itemize}

\begin{itemize}
    \setlength\itemsep{0em}
    \item \textbf{Input.} our orangutans who do swim \textbf{don't} eat.
    \item \textbf{Output.} \textbf{don't} our orangutans who do swim eat?
\end{itemize}

\subsection{MultiNLI Examples}
\begin{itemize}
    \setlength\itemsep{0em}
    \item \textbf{Premise.} Conceptually cream skimming has two basic dimensions - product and geography.
    \item \textbf{Hypothesis.} Product and geography are what make cream skimming work.
    \item \textbf{Label.} \textbf{Neutral}
\end{itemize}

\begin{itemize}
    \setlength\itemsep{0em}
    \item \textbf{Premise.} How do you know? All this is their information again.
    \item \textbf{Hypothesis.} This information belongs to them.
    \item \textbf{Label.} \textbf{Entailment}
\end{itemize}

\begin{itemize}
    \setlength\itemsep{0em}
    \item \textbf{Premise.} Fun for adults and children.
    \item \textbf{Hypothesis.} Fun for only children.
    \item \textbf{Label.} \textbf{Contradiction}
\end{itemize}

\subsection{MoNLI Examples}
\begin{itemize}
    \setlength\itemsep{0em}
    \item \textbf{Premise.} The man does not own a dog.
    \item \textbf{Hypothesis.} The man does not own a mammal.
    \item \textbf{Label.} \textbf{Neutral}
\end{itemize}

\begin{itemize}
    \setlength\itemsep{0em}
    \item \textbf{Premise.} The man does not own a mammal.
    \item \textbf{Hypothesis.} The man does not own a dog.
    \item \textbf{Label.} \textbf{Entailment}
\end{itemize}

\subsection{MED Examples}
\begin{itemize}
    \setlength\itemsep{0em}
    \item \textbf{Premise.} No delegate finished the report on time.
    \item \textbf{Hypothesis.} No delegate finished the report.
    \item \textbf{Label.} \textbf{Neutral}
\end{itemize}

\begin{itemize}
    \setlength\itemsep{0em}
    \item \textbf{Premise.} Some delegates finished the survey on time.
    \item \textbf{Hypothesis.} Some delegates finished the survey.
    \item \textbf{Label.} \textbf{Entailment}
\end{itemize}

\section{NLI Setup}
\label{sec:nliapp}

We now discuss the training and evaluation setup used for our NLI experiments. We format the training examples using the following prompt:

\begin{quote}
Determine if the hypothesis is an entailment, contradiction or neutral in relation to the premise. If the hypothesis directly follows from the premise, it is an entailment. If the hypothesis directly contradicts the premise, it is a contradiction. Any relationship that does not fit in with the above definitions is considered neutral. Return A if the second sentence is an entailment, B if it is a contradiction, and C if it is neutral.

\#\#\# Premise

\{premise\}

\#\#\# Hypothesis

\{hypothesis\}

\#\#\# Answer
\end{quote}

We frame training as a classification task on the final logit from the LLM when the prompt above is passed as input. Instead of using a classification head, we directly apply cross-entropy loss over the log-probabilities of 'A', 'B', and 'C' as possible continuations of the prompt, with the groundtruth class label as the target. During evaluation, we select the continuation with the highest log-probability at the final logit, with 'A', 'B' and 'C' as choices. Note that we train on MultiNLI, which includes examples of the entailment, contradiction and neutral labels. On the other hand, MoNLI and MED datasets treat NLI as a two-class problem (entailment or neutral). However, we retain all three classes as options during inference on these datasets, for consistency with training.

\section{Experiment Hyperparameters}
\label{sec:hyperparams}
In this section, we detail the hyperparameters used for all of our main experiments. All experiments with LMs trained from scratch use tied input and output weight matrices. For all experiments, linear warmup is used for learning rates for the first 10\% of training steps, after which cosine decay is used to reach a learning rate of 0 at the end of training. We use the AdamW optimizer with epsilon of 1e-7 and clip gradients to a maximum $L_2$ norm of 1.0.

\subsection{\S~\ref{sec:grokking}}
\begin{itemize}
    \setlength\itemsep{0em}
    \item Number of layers: 4
    \item Number of attention heads: 8
    \item Hidden dimension: 512
    \item Training steps: 500,000
    \item Batch size: 8
    \item Learning rate: 1e-4
    \item Weight decay: 0.01
    \item Dropout: 0.1
\end{itemize}

\subsection{\S~\ref{sec:bllip}}
\begin{itemize}
    \setlength\itemsep{0em}
    \item Number of layers: 16
    \item Number of attention heads: 8
    \item Hidden dimension: 512
    \item Training steps: 100,000
    \item Batch size: 160
    \item Learning rate: 1e-4
    \item Weight decay: 0.01
    \item Dropout: 0.1
\end{itemize}

\subsection{\S~\ref{sec:wikitext}}
\begin{itemize}
    \setlength\itemsep{0em}
    \item Number of layers: 12
    \item Number of attention heads: 12
    \item Hidden dimension: 768
    \item Training steps: 40,000
    \item Batch size: 480
    \item Learning rate: 6e-4
    \item Weight decay: 0.1
    \item Dropout: 0.1
\end{itemize}

\subsection{\S~\ref{sec:cpt}}
\begin{itemize}
    \setlength\itemsep{0em}
    \item Training steps: 10,000
    \item Batch size: 32
    \item Learning rate: 2e-5
    \item Weight decay: 0.01
    \item Dropout: 0.1
\end{itemize}

\subsection{\S~\ref{sec: NLI}}
\begin{itemize}
    \setlength\itemsep{0em}
    \item Training steps: 20,000
    \item Batch size: 32
    \item Learning rate: 1e-5
    \item Weight decay: 0.01
    \item Dropout: 0.1
\end{itemize}

\section{Selection of Layer and Number of Attention Heads for $\ours{}$}
\label{sec:ablation}

\begin{figure}[h]
\setlength\belowcaptionskip{-5pt}
    \centering
    \includegraphics[width=0.5\textwidth]{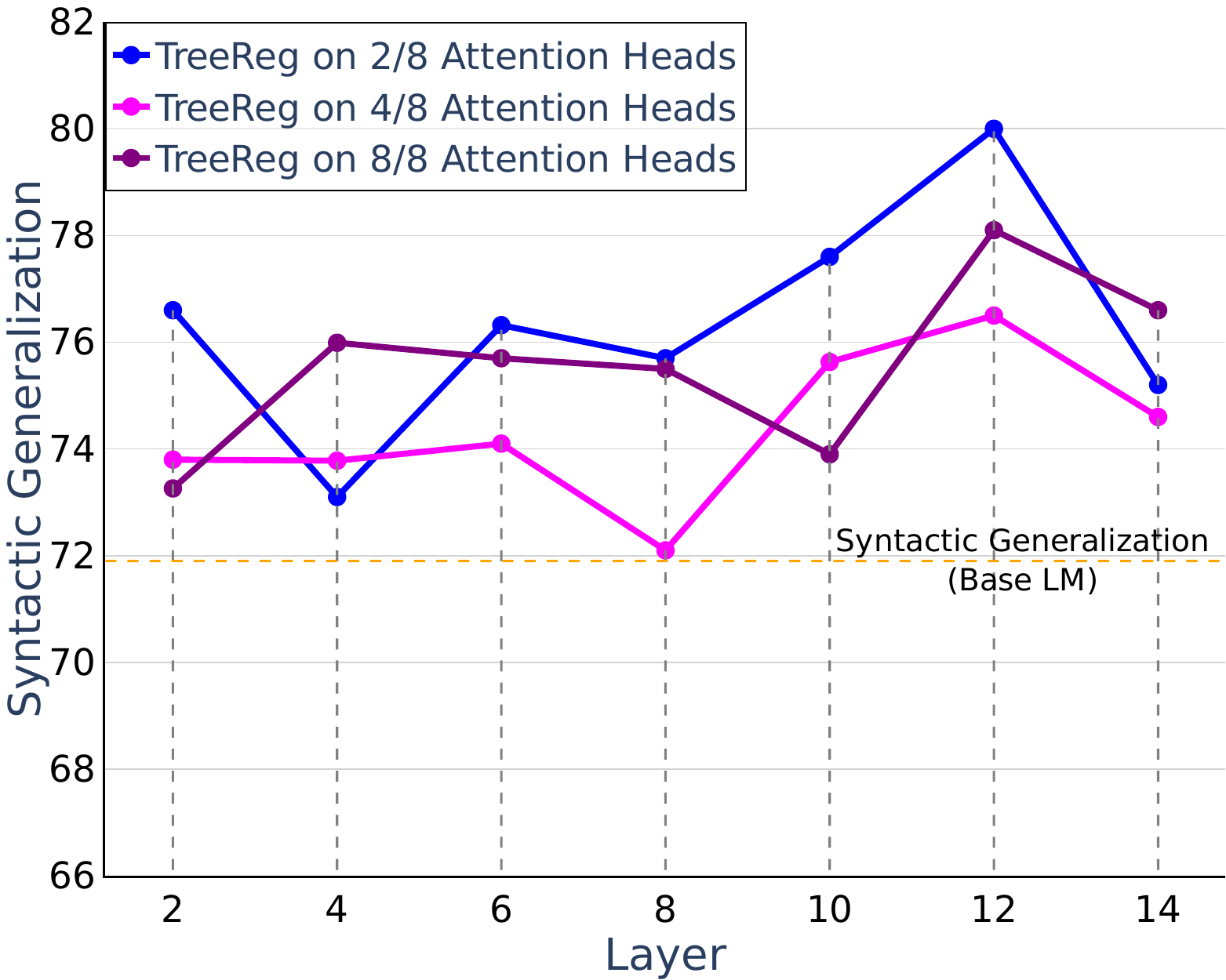}\vspace{-0.05in}
    \captionsetup{skip=10pt}
    \caption{Syntactic Generalization on SG test suites for different layers and number of attention heads used in $\ours{}$, for LMs trained from scratch on BLLIP-LG.}
    \label{fig:sgabl}
\end{figure}

\begin{figure}[h]
\setlength\belowcaptionskip{-5pt}
    \centering
    \includegraphics[width=0.5\textwidth]{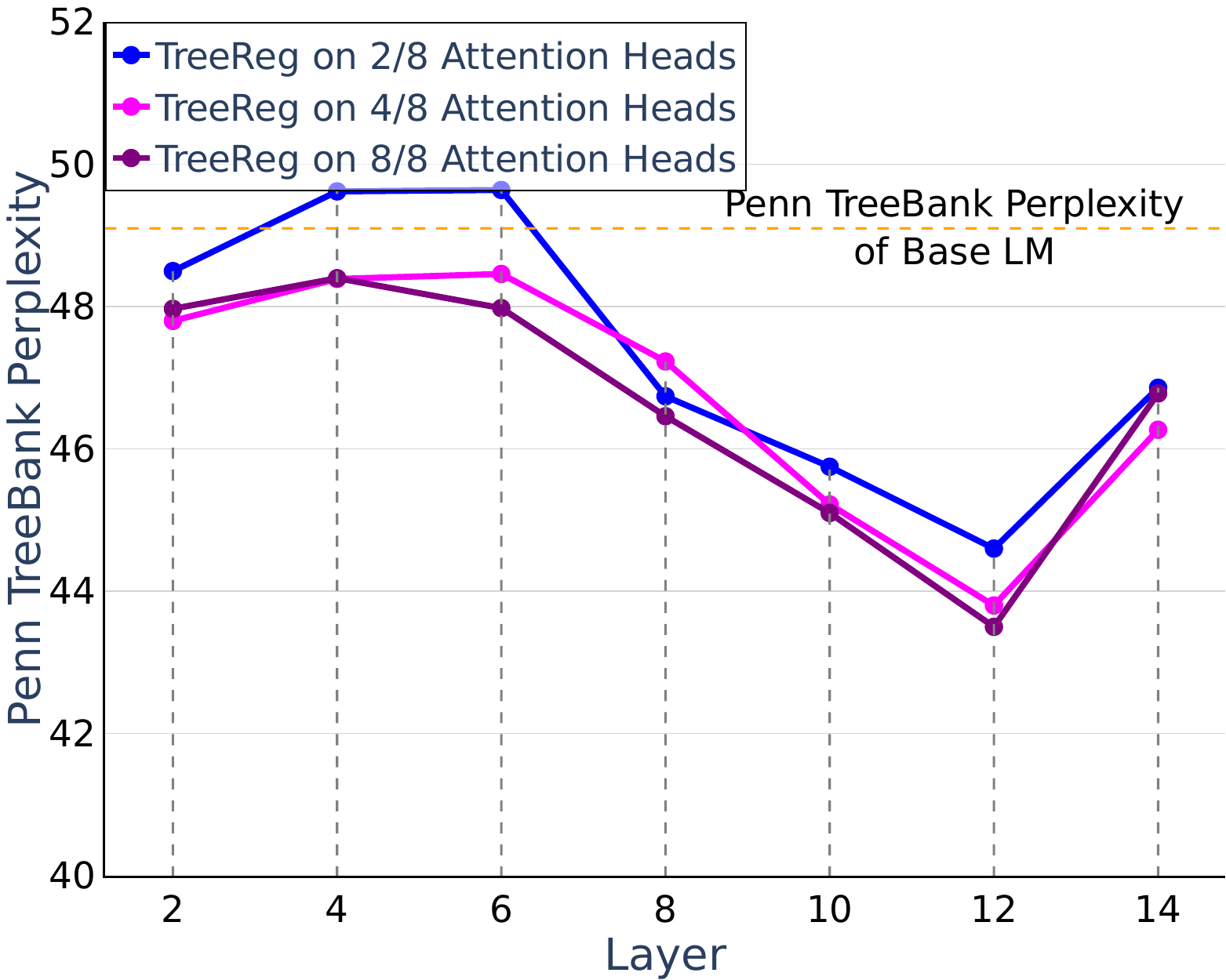}\vspace{-0.05in}
    \captionsetup{skip=10pt}
    \caption{PTB Perplexity on SG test suites for different layers and number of attention heads used in $\ours{}$, for LMs trained from scratch on BLLIP-LG.}
    \label{fig:ptbabl}
\end{figure}

What is the ideal layer and number of attention heads for the $\LMTR{}$ computation? To explore, we vary the layer used in $\ours{}$ as [2,4,6,8,10,12,14] and the number of attention heads used as [2,4,8] out of 8 in the experiment setup described in Section~\ref{sec:bllip} for the training of a 16-layer LM from scratch on the BLLIP-LG dataset. For the resulting models, we report the overall SyntaxGym performance in Figure~\ref{fig:sgabl} and PTB perplexities in Figure~\ref{fig:ptbabl}. 

 We find that applying $\ours{}$ at layer 12 on 2 attention heads results in the highest overall SyntaxGym performance as well as lowest PTB perplexity. With a couple of exceptions (4 
 attention heads at layer 4 and 6), $\ours{}$ consistently outperforms the Base LM in both SyntaxGym performance and PTB perplexities across hyperparameter settings. Both syntactic generalization and out-of-distribution perplexities show generally improving trends till layer 12, after which they start degrading. This finding suggests that $\ours{}$ LMs represent syntax most naturally at intermediate layers, on a small subset of attention heads.

 % Pareto front between SG and PPL, interpret those
 % Rerun the crashed runs (l10 0.25 etc)
% move this up
 % too controversial? is there something similar in literature?

\section{Examples of Parses Induced by $\ours{}$}
\label{sec:parses}

In this section, we provide some examples of parses induced from the $\ours{}$ LM trained in \S\ref{sec:bllip} on the BLLIP-LG, PTB and 4000 Questions test sets.

\subsection{BLLIP-LG}
\begin{figure}[h]
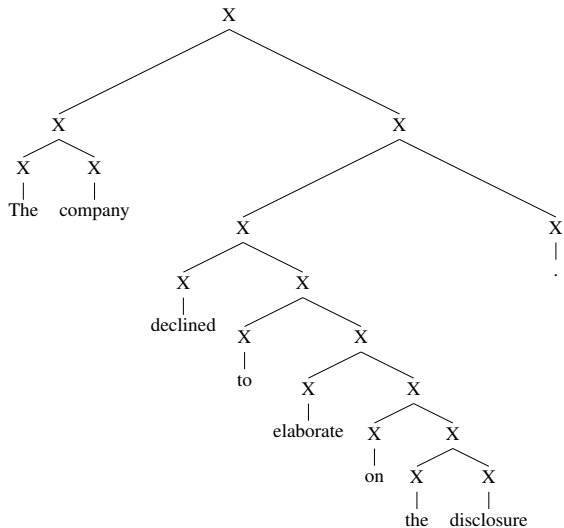

\setlength\belowcaptionskip{-5pt}
    \centering
    \resizebox{\columnwidth}{!}{
    \begin{tabular}{c}
    \Tree [.X        [.X [.X The ] [.X company ] ]        [.X         [.X            [.X declined ]            [.X              [.X to ]              [.X                [.X elaborate ]
       [.X [.X on ] [.X [.X the ] [.X disclosure ] ] ] ] ] ]          [.X . ] ] ]
    \end{tabular}
    }
    \captionsetup{skip=10pt}
    \caption{Silver and induced parse for ``The company declined to elaborate on the disclosure.''}
    \label{fig:b_p_1}
\end{figure}

\begin{figure}[h]
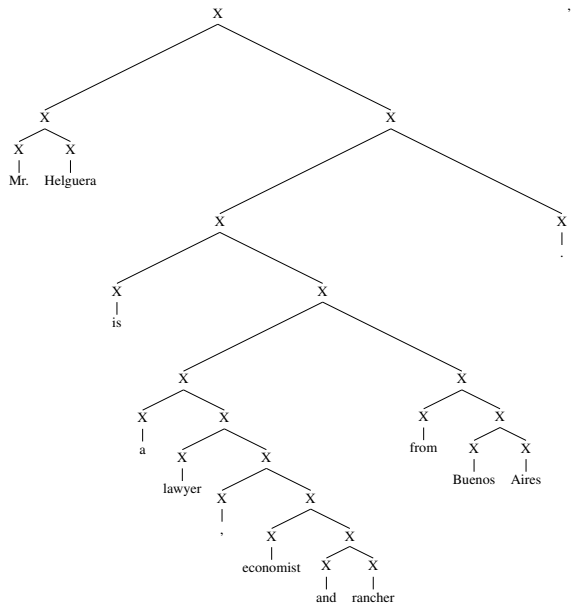

\setlength\belowcaptionskip{-5pt}
    \centering
    \resizebox{\columnwidth}{!}{
    \begin{tabular}{c}
    \Tree [.X        [.X [.X Mr. ] [.X Helguera ] ]        [.X          [.X            [.X is ]            [.X              [.X                [.X a ]                [.X                  [.X lawyer ]                  [.X                    [.X , ]
         [.X                      [.X economist ]
     [.X [.X and ] [.X rancher ] ] ] ] ] ]              [.X [.X from ] [.X [.X Buenos ] [.X Aires ] ] ] ] ]          [.X . ] ] ]'
    \end{tabular}
    }
    \captionsetup{skip=10pt}
    \caption{Silver parse for ``Mr. Helguera is a lawyer, economist and rancher from Buenos Aires.''}
    \label{fig:b_p_2}
\end{figure}

\begin{figure}[h]
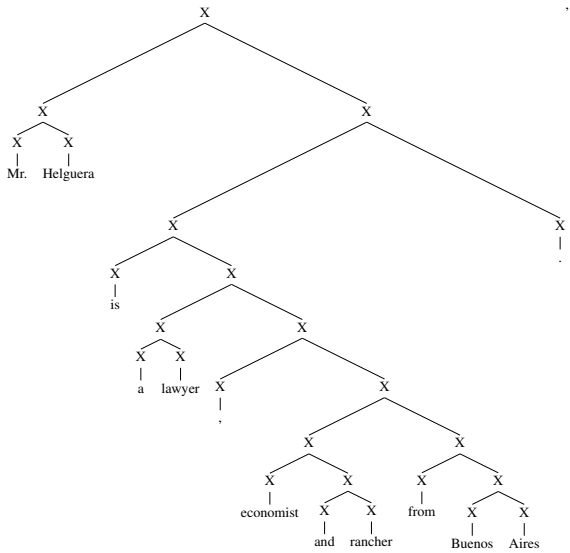

\setlength\belowcaptionskip{-5pt}
    \centering
    \resizebox{\columnwidth}{!}{
    \begin{tabular}{c}
    \Tree [.X        [.X [.X Mr. ] [.X Helguera ] ]        [.X          [.X            [.X is ]            [.X              [.X [.X a ] [.X lawyer ] ]              [.X                [.X , ]                [.X                  [.X [.X economist ] [.X [.X and ] [.X rancher ] ] ]                  [.X [.X from ] [.X [.X Buenos ] [.X Aires ] ] ] ] ] ] ]          [.X . ] ] ]'
    \end{tabular}
    }
    \captionsetup{skip=10pt}
    \caption{Induced parse for ``Mr. Helguera is a lawyer, economist and rancher from Buenos Aires.''}
    \label{fig:b_p_3}
\end{figure}

Figure~\ref{fig:b_p_1} is an example where the induced parse from $\ours{}$ matches the silver parse. Figure~\ref{fig:b_p_2} and Figure~\ref{fig:b_p_3} represents a case where the induced parse differs from the silver parse.

\subsection{PTB}
\begin{figure}[h]
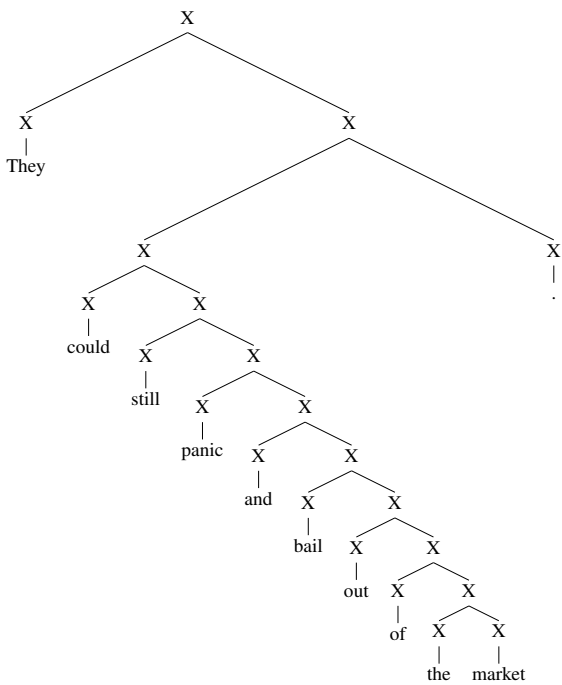

\setlength\belowcaptionskip{-5pt}
    \centering
    \resizebox{\columnwidth}{!}{
    \begin{tabular}{c}
    \Tree [.X        [.X They ]        [.X          [.X
   [.X could ]            [.X              [.X still ]
    [.X                [.X panic ]                [.X
       [.X and ]                  [.X                    [.X bail ]                    [.X                      [.X out ]
             [.X [.X of ] [.X [.X the ] [.X market ] ] ] ] ] ] ] ] ]          [.X . ] ] ]
    \end{tabular}
    }
    \captionsetup{skip=10pt}
    \caption{Silver and induced parse for ``They could still panic and bail out of the market.''}
    \label{fig:p_p_1}
\end{figure}

\begin{figure}[h]
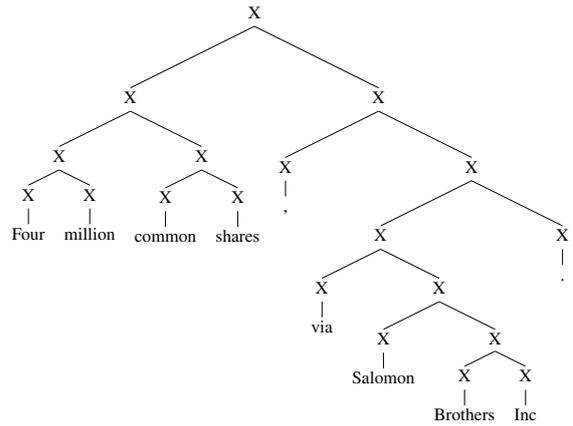

\setlength\belowcaptionskip{-5pt}
    \centering
    \resizebox{\columnwidth}{!}{
    \begin{tabular}{c}
    \Tree [.X        [.X          [.X [.X Four ] [.X million ] ]          [.X [.X common ] [.X shares ] ] ]        [.X          [.X , ]          [.X            [.X              [.X via ]              [.X [.X Salomon ] [.X [.X Brothers ] [.X Inc ] ] ] ]
  [.X . ] ] ] ]
    \end{tabular}
    }
    \captionsetup{skip=10pt}
    \caption{Silver parse for `Four million common shares, via Salomon Brothers Inc.''}
    \label{fig:p_p_2}
\end{figure}

\begin{figure}[h]
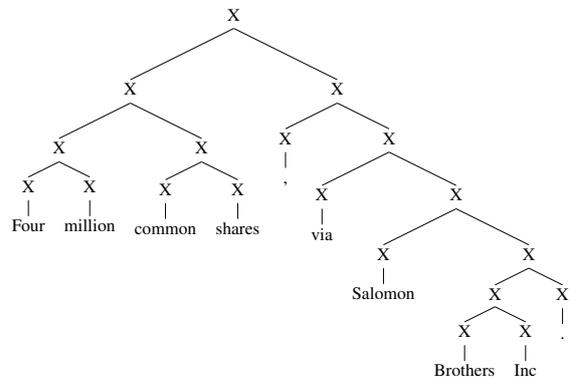

\setlength\belowcaptionskip{-5pt}
    \centering
    \resizebox{\columnwidth}{!}{
    \begin{tabular}{c}
    \Tree [.X        [.X          [.X [.X Four ] [.X million ] ]          [.X [.X common ] [.X shares ] ] ]        [.X          [.X , ]          [.X            [.X via ]            [.X
    [.X Salomon ]              [.X [.X [.X Brothers ] [.X Inc ] ] [.X . ] ] ] ] ] ]
    \end{tabular}
    }
    \captionsetup{skip=10pt}
    \caption{Induced parse for ``Four million common shares, via Salomon Brothers Inc.''}
    \label{fig:p_p_3}
\end{figure}

Figure~\ref{fig:p_p_1} is an example where the induced parse from $\ours{}$ matches the silver parse. Figure~\ref{fig:p_p_2} and Figure~\ref{fig:p_p_3} represents a case where the induced parse differs from the silver parse.

\subsection{4000 Questions}
\begin{figure}[h]
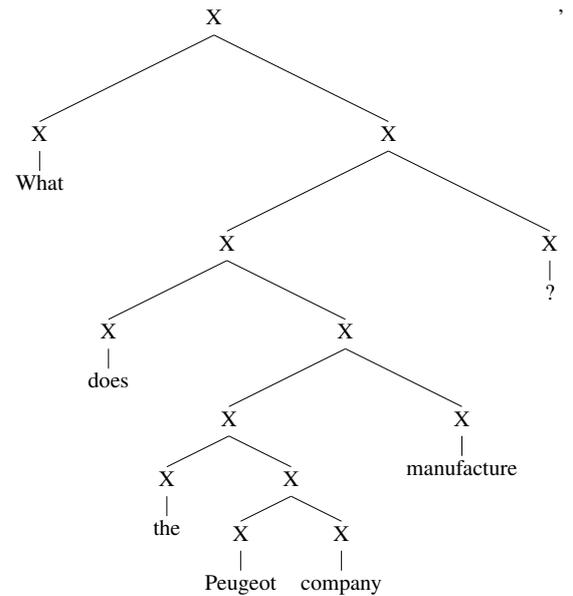

\setlength\belowcaptionskip{-5pt}
    \centering
    \resizebox{\columnwidth}{!}{
    \begin{tabular}{c}
    \Tree [.X        [.X What ]        [.X          [.X
   [.X does ]            [.X              [.X [.X the ] [.X [.X Peugeot ] [.X company ] ] ]              [.X manufacture ] ] ]          [.X ? ] ] ]'
    \end{tabular}
    }
    \captionsetup{skip=10pt}
    \caption{Silver and induced parse for ``What does the Peugeot company manufacture?''}
    \label{fig:4_p_1}
\end{figure}

\begin{figure}[h]
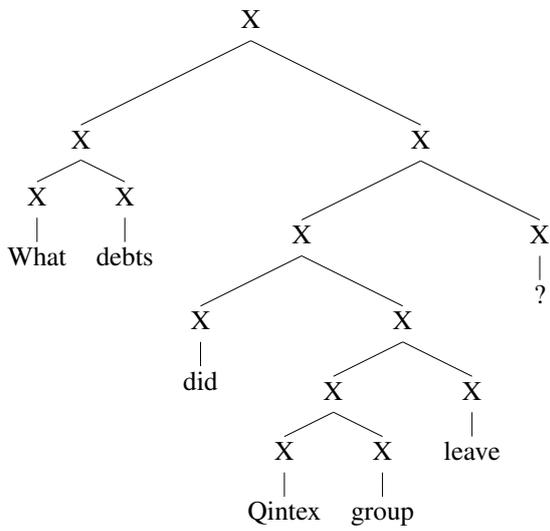

\setlength\belowcaptionskip{-5pt}
    \centering
    \resizebox{\columnwidth}{!}{
    \begin{tabular}{c}
    \Tree [.X        [.X [.X What ] [.X debts ] ]        [.X          [.X            [.X did ]            [.X [.X [.X Qintex ] [.X group ] ] [.X leave ] ] ]          [.X ? ] ] ]
    \end{tabular}
    }
    \captionsetup{skip=10pt}
    \caption{Silver parse for ``What debts did Qintex group leave?''}
    \label{fig:4_p_2}
\end{figure}

\begin{figure}[h]
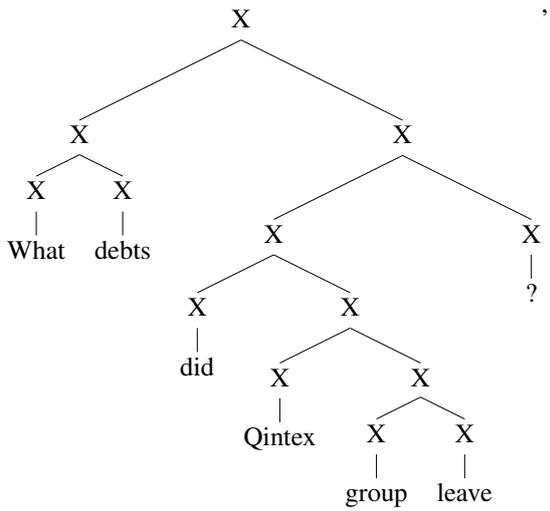

\setlength\belowcaptionskip{-5pt}
    \centering
    \resizebox{\columnwidth}{!}{
    \begin{tabular}{c}
    \Tree [.X        [.X [.X What ] [.X debts ] ]        [.X          [.X            [.X did ]            [.X [.X Qintex ] [.X [.X group ] [.X leave ] ] ] ]          [.X ? ] ] ]'
    \end{tabular}
    }
    \captionsetup{skip=10pt}
    \caption{Induced parse for ``What debts did Qintex group leave?''}
    \label{fig:4_p_3}
\end{figure}

Figure~\ref{fig:4_p_1} is an example where the induced parse from $\ours{}$ matches the silver parse. Figure~\ref{fig:4_p_2} and Figure~\ref{fig:4_p_3} represents a case where the induced parse differs from the silver parse.

\end{document}

%% file: macros.tex
\newcommand{\hrow}{\rowcolor{gray!15}}

\newcommand{\ours}{\textsc{TreeReg}}
\newcommand{\SCIS}{\mathrm{SCIN}}
\newcommand{\splitscore}{s}
\newcommand{\LMLoss}{\ensuremath{\mathcal{L}_\text{LM}}}
\newcommand{\LMTR}{\ensuremath{\mathcal{L}_\text{TR}}}
\newcommand{\range}[2]{S_{#1; #2}}
\newcommand{\uprm}[1]{\ensuremath{^{\hbox{\scriptsize #1}}}}
\newcommand{\cvec}[1]{\ensuremath{\bm{#1}}}
\DeclareMathOperator*{\argmax}{arg\,max}
\newcommand{\cmark}{\ding{51}} % Check mark
\newcommand{\xmark}{\ding{55}} % Cross mark

\newcommand{\gparse}[1]{\ensuremath{\hat{T}(#1)}}

%% file: main.bbl
\begin{thebibliography}{51}
\providecommand{\natexlab}[1]{#1}

\bibitem[{Achiam et~al.(2023)Achiam, Adler, Agarwal, Ahmad, Akkaya, Aleman, Almeida, Altenschmidt, Altman, Anadkat et~al.}]{achiam2023gpt}
Josh Achiam, Steven Adler, Sandhini Agarwal, Lama Ahmad, Ilge Akkaya, Florencia~Leoni Aleman, Diogo Almeida, Janko Altenschmidt, Sam Altman, Shyamal Anadkat, et~al. 2023.
\newblock Gpt-4 technical report.
\newblock \emph{arXiv preprint arXiv:2303.08774}.

\bibitem[{Allen-Zhu and Li(2023)}]{allen2023physics}
Zeyuan Allen-Zhu and Yuanzhi Li. 2023.
\newblock Physics of language models: Part 1, learning hierarchical language structures.
\newblock \emph{arXiv preprint arXiv:2305.13673}.

\bibitem[{Berglund et~al.(2024)Berglund, Tong, Kaufmann, Balesni, Stickland, Korbak, and Evans}]{berglund2023reversal}
Lukas Berglund, Meg Tong, Maximilian Kaufmann, Mikita Balesni, Asa~Cooper Stickland, Tomasz Korbak, and Owain Evans. 2024.
\newblock The reversal curse: {LLMs} trained on ``{A} is {B}'' fail to learn ``{B} is {A}''.
\newblock In \emph{Proceedings of the International Conference on Learning Representations}.

\bibitem[{Bhattamishra et~al.(2020)Bhattamishra, Ahuja, and Goyal}]{bhattamishra-etal-2020-ability}
Satwik Bhattamishra, Kabir Ahuja, and Navin Goyal. 2020.
\newblock On the {A}bility and {L}imitations of {T}ransformers to {R}ecognize {F}ormal {L}anguages.
\newblock In \emph{Proceedings of the Conference on Empirical Methods in Natural Language Processing}.

\bibitem[{Brown et~al.(2020)Brown, Mann, Ryder, Subbiah, Kaplan, Dhariwal, Neelakantan, Shyam, Sastry, Askell, Agarwal, Herbert-Voss, Krueger, Henighan, Child, Ramesh, Ziegler, Wu, Winter, Hesse, Chen, Sigler, Litwin, Gray, Chess, Clark, Berner, McCandlish, Radford, Sutskever, and Amodei}]{brown2020language}
Tom~B. Brown, Benjamin Mann, Nick Ryder, Melanie Subbiah, Jared Kaplan, Prafulla Dhariwal, Arvind Neelakantan, Pranav Shyam, Girish Sastry, Amanda Askell, Sandhini Agarwal, Ariel Herbert-Voss, Gretchen Krueger, Tom Henighan, Rewon Child, Aditya Ramesh, Daniel~M. Ziegler, Jeffrey Wu, Clemens Winter, Christopher Hesse, Mark Chen, Eric Sigler, Mateusz Litwin, Scott Gray, Benjamin Chess, Jack Clark, Christopher Berner, Sam McCandlish, Alec Radford, Ilya Sutskever, and Dario Amodei. 2020.
\newblock Language models are few-shot learners.
\newblock In \emph{Proceedings of the International Conference on Neural Information Processing Systems}.

\bibitem[{Chiang and Cholak(2022)}]{chiang-cholak-2022-overcoming}
David Chiang and Peter Cholak. 2022.
\newblock Overcoming a theoretical limitation of self-attention.
\newblock In \emph{Proceedings of the Annual Meeting of the Association for Computational Linguistics (Volume 1: Long Papers)}.

\bibitem[{Cocke(1969)}]{cocke1969programming}
John Cocke. 1969.
\newblock \emph{Programming languages and their compilers: Preliminary notes}.
\newblock New York University.

\bibitem[{Crain and Nakayama(1987)}]{crain1987structure}
Stephen Crain and Mineharu Nakayama. 1987.
\newblock Structure dependence in grammar formation.
\newblock \emph{Language}, pages 522--543.

\bibitem[{Deshpande and Narasimhan(2020)}]{deshpande-narasimhan-2020-guiding}
Ameet Deshpande and Karthik Narasimhan. 2020.
\newblock Guiding attention for self-supervised learning with transformers.
\newblock In \emph{Findings of the Association for Computational Linguistics: EMNLP}.

\bibitem[{Dubey et~al.(2024)Dubey, Jauhri, Pandey, Kadian, Al-Dahle, Letman, Mathur, Schelten, Yang, Fan et~al.}]{dubey2024llama}
Abhimanyu Dubey, Abhinav Jauhri, Abhinav Pandey, Abhishek Kadian, Ahmad Al-Dahle, Aiesha Letman, Akhil Mathur, Alan Schelten, Amy Yang, Angela Fan, et~al. 2024.
\newblock The llama 3 herd of models.
\newblock \emph{arXiv preprint arXiv:2407.21783}.

\bibitem[{Dyer et~al.(2016)Dyer, Kuncoro, Ballesteros, and Smith}]{dyer-etal-2016-recurrent}
Chris Dyer, Adhiguna Kuncoro, Miguel Ballesteros, and Noah~A. Smith. 2016.
\newblock Recurrent neural network grammars.
\newblock In \emph{Proceedings of the Conference of the North {A}merican Chapter of the Association for Computational Linguistics: Human Language Technologies}.

\bibitem[{Dziri et~al.(2023)Dziri, Lu, Sclar, Li, Jiang, Lin, Welleck, West, Bhagavatula, Le~Bras, Hwang, Sanyal, Ren, Ettinger, Harchaoui, and Choi}]{NEURIPS2023_deb3c281}
Nouha Dziri, Ximing Lu, Melanie Sclar, Xiang~(Lorraine) Li, Liwei Jiang, Bill~Yuchen Lin, Sean Welleck, Peter West, Chandra Bhagavatula, Ronan Le~Bras, Jena Hwang, Soumya Sanyal, Xiang Ren, Allyson Ettinger, Zaid Harchaoui, and Yejin Choi. 2023.
\newblock Faith and fate: Limits of transformers on compositionality.
\newblock In \emph{Proceedings of the International Conference in Neural Information Processing Systems}.

\bibitem[{Eriguchi et~al.(2016)Eriguchi, Hashimoto, and Tsuruoka}]{eriguchi-etal-2016-tree}
Akiko Eriguchi, Kazuma Hashimoto, and Yoshimasa Tsuruoka. 2016.
\newblock Tree-to-sequence attentional neural machine translation.
\newblock In \emph{Proceedings of the Annual Meeting of the Association for Computational Linguistics (Volume 1: Long Papers)}.

\bibitem[{Gauthier et~al.(2020)Gauthier, Hu, Wilcox, Qian, and Levy}]{gauthier2020syntaxgym}
Jon Gauthier, Jennifer Hu, Ethan Wilcox, Peng Qian, and Roger Levy. 2020.
\newblock Syntaxgym: An online platform for targeted evaluation of language models.
\newblock In \emph{Proceedings of the Annual Meeting of the Association for Computational Linguistics: System Demonstrations}.

\bibitem[{Geiger et~al.(2020)Geiger, Richardson, and Potts}]{geiger2020neural}
Atticus Geiger, Kyle Richardson, and Christopher Potts. 2020.
\newblock Neural natural language inference models partially embed theories of lexical entailment and negation.
\newblock In \emph{Proceedings of the BlackboxNLP Workshop on Analyzing and Interpreting Neural Networks for NLP}.

\bibitem[{Guo et~al.(2020)Guo, Lin, Lou, and Zhang}]{guo2020hierarchical}
Yinuo Guo, Zeqi Lin, Jian-Guang Lou, and Dongmei Zhang. 2020.
\newblock Hierarchical poset decoding for compositional generalization in language.
\newblock In \emph{Proceedings of the International Conference on Neural Information Processing Systems}.

\bibitem[{Hahn(2020)}]{hahn2020theoretical}
Michael Hahn. 2020.
\newblock Theoretical limitations of self-attention in neural sequence models.
\newblock \emph{Transactions of the Association for Computational Linguistics}, 8:156--171.

\bibitem[{Hale et~al.(2018)Hale, Dyer, Kuncoro, and Brennan}]{hale2018finding}
John Hale, Chris Dyer, Adhiguna Kuncoro, and Jonathan Brennan. 2018.
\newblock Finding syntax in human encephalography with beam search.
\newblock In \emph{Proceedings of the Annual Meeting of the Association for Computational Linguistics (Volume 1: Long Papers)}.

\bibitem[{Hale and Stanojevi{\'c}(2024)}]{hale-stanojevic-2024-llms}
John~T. Hale and Milo{\v{s}} Stanojevi{\'c}. 2024.
\newblock Do {LLM}s learn a true syntactic universal?
\newblock In \emph{Proceedings of the Conference on Empirical Methods in Natural Language Processing}.

\bibitem[{Hu et~al.(2020)Hu, Gauthier, Qian, Wilcox, and Levy}]{hu-etal-2020-systematic}
Jennifer Hu, Jon Gauthier, Peng Qian, Ethan Wilcox, and Roger Levy. 2020.
\newblock A systematic assessment of syntactic generalization in neural language models.
\newblock In \emph{Proceedings of the Annual Meeting of the Association for Computational Linguistics}.

\bibitem[{Hu et~al.(2024)Hu, Ji, Zhu, Wu, and Tu}]{hu2024generative}
Xiang Hu, Pengyu Ji, Qingyang Zhu, Wei Wu, and Kewei Tu. 2024.
\newblock Generative pretrained structured transformers: Unsupervised syntactic language models at scale.
\newblock \emph{arXiv preprint arXiv:2403.08293}.

\bibitem[{Hu et~al.(2021)Hu, Mi, Wen, Wang, Su, Zheng, and de~Melo}]{hu-etal-2021-r2d2}
Xiang Hu, Haitao Mi, Zujie Wen, Yafang Wang, Yi~Su, Jing Zheng, and Gerard de~Melo. 2021.
\newblock {R}2{D}2: Recursive transformer based on differentiable tree for interpretable hierarchical language modeling.
\newblock In \emph{Proceedings of the Annual Meeting of the Association for Computational Linguistics and the International Joint Conference on Natural Language Processing (Volume 1: Long Papers)}.

\bibitem[{Judge et~al.(2006)Judge, Cahill, and van Genabith}]{judge2006questionbank}
John Judge, Aoife Cahill, and Josef van Genabith. 2006.
\newblock {Q}uestion{B}ank: Creating a corpus of parse-annotated questions.
\newblock In \emph{Proceedings of the International Conference on Computational Linguistics and Annual Meeting of the Association for Computational Linguistics}.

\bibitem[{Kasami(1966)}]{kasami1966efficient}
Tadao Kasami. 1966.
\newblock An efficient recognition and syntax-analysis algorithm for context-free languages.
\newblock \emph{Coordinated Science Laboratory Report no. R-257}.

\bibitem[{Kitaev et~al.(2022)Kitaev, Lu, and Klein}]{kitaev2022learned}
Nikita Kitaev, Thomas Lu, and Dan Klein. 2022.
\newblock Learned incremental representations for parsing.
\newblock In \emph{Proceedings of the Annual Meeting of the Association for Computational Linguistics (Volume 1: Long Papers)}.

\bibitem[{Le and Zuidema(2015)}]{le-zuidema-2015-compositional}
Phong Le and Willem Zuidema. 2015.
\newblock Compositional distributional semantics with long short term memory.
\newblock In \emph{Proceedings of the Joint Conference on Lexical and Computational Semantics}.

\bibitem[{Marcus et~al.(1993)Marcus, Santorini, and Marcinkiewicz}]{marcus1993building}
Mitch Marcus, Beatrice Santorini, and Mary~Ann Marcinkiewicz. 1993.
\newblock Building a large annotated corpus of english: The penn treebank.
\newblock \emph{Computational linguistics}, 19(2):313--330.

\bibitem[{McCoy et~al.(2020)McCoy, Frank, and Linzen}]{mccoy2020does}
R~Thomas McCoy, Robert Frank, and Tal Linzen. 2020.
\newblock Does syntax need to grow on trees? sources of hierarchical inductive bias in sequence-to-sequence networks.
\newblock \emph{Transactions of the Association for Computational Linguistics}, 8:125--140.

\bibitem[{McCoy et~al.(2019)McCoy, Pavlick, and Linzen}]{mccoy2019right}
Tom McCoy, Ellie Pavlick, and Tal Linzen. 2019.
\newblock Right for the wrong reasons: Diagnosing syntactic heuristics in natural language inference.
\newblock In \emph{Proceedings of the Annual Meeting of the Association for Computational Linguistics}.

\bibitem[{Merity et~al.(2022)Merity, Xiong, Bradbury, and Socher}]{merity2022pointer}
Stephen Merity, Caiming Xiong, James Bradbury, and Richard Socher. 2022.
\newblock Pointer sentinel mixture models.
\newblock In \emph{Proceedings of the International Conference on Learning Representations}.

\bibitem[{Murty et~al.(2023{\natexlab{a}})Murty, Sharma, Andreas, and Manning}]{murty2023grokking}
Shikhar Murty, Pratyusha Sharma, Jacob Andreas, and Christopher Manning. 2023{\natexlab{a}}.
\newblock Grokking of hierarchical structure in vanilla transformers.
\newblock In \emph{Proceedings of the Annual Meeting of the Association for Computational Linguistics (Volume 2: Short Papers)}.

\bibitem[{Murty et~al.(2023{\natexlab{b}})Murty, Sharma, Andreas, and Manning}]{murty2023pushdown}
Shikhar Murty, Pratyusha Sharma, Jacob Andreas, and Christopher Manning. 2023{\natexlab{b}}.
\newblock Pushdown layers: Encoding recursive structure in transformer language models.
\newblock In \emph{Proceedings of the Conference on Empirical Methods in Natural Language Processing}.

\bibitem[{Murty et~al.(2023{\natexlab{c}})Murty, Sharma, Andreas, and Manning}]{murty2022characterizing}
Shikhar Murty, Pratyusha Sharma, Jacob Andreas, and Christopher~D Manning. 2023{\natexlab{c}}.
\newblock Characterizing intrinsic compositionality in transformers with tree projections.
\newblock In \emph{Proceedings of the International Conference on Learning Representations}.

\bibitem[{Pallier et~al.(2011)Pallier, Devauchelle, and Dehaene}]{pallier2011cortical}
Christophe Pallier, Anne-Dominique Devauchelle, and Stanislas Dehaene. 2011.
\newblock Cortical representation of the constituent structure of sentences.
\newblock \emph{Proceedings of the National Academy of Sciences}, 108(6):2522--2527.

\bibitem[{Qian et~al.(2021)Qian, Naseem, Levy, and Astudillo}]{qian2021structural}
Peng Qian, Tahira Naseem, Roger Levy, and Ram{\'o}n~Fernandez Astudillo. 2021.
\newblock Structural guidance for transformer language models.
\newblock In \emph{Proceedings of the Annual Meeting of the Association for Computational Linguistics and the International Joint Conference on Natural Language Processing (Volume 1: Long Papers)}.

\bibitem[{Sartran et~al.(2022)Sartran, Barrett, Kuncoro, Stanojevi{\'c}, Blunsom, and Dyer}]{sartran2022transformer}
Laurent Sartran, Samuel Barrett, Adhiguna Kuncoro, Milo{\v{s}} Stanojevi{\'c}, Phil Blunsom, and Chris Dyer. 2022.
\newblock Transformer grammars: Augmenting transformer language models with syntactic inductive biases at scale.
\newblock \emph{Transactions of the Association for Computational Linguistics}, 10:1423--1439.

\bibitem[{Sheehan et~al.(2017)Sheehan, Biberauer, Roberts, and Holmberg}]{sheehan2017final}
Michelle Sheehan, Theresa Biberauer, Ian Roberts, and Anders Holmberg. 2017.
\newblock \emph{The Final-Over-Final Condition: A Syntactic Universal}, volume~76.
\newblock MIT Press.

\bibitem[{Shen et~al.(2019)Shen, Tan, Sordoni, and Courville}]{shen2018ordered}
Yikang Shen, Shawn Tan, Alessandro Sordoni, and Aaron Courville. 2019.
\newblock Ordered neurons: Integrating tree structures into recurrent neural networks.
\newblock In \emph{Proceedings of the International Conference on Learning Representations}.

\bibitem[{Socher et~al.(2013)Socher, Perelygin, Wu, Chuang, Manning, Ng, and Potts}]{socher2013recursive}
Richard Socher, Alex Perelygin, Jean Wu, Jason Chuang, Christopher~D Manning, Andrew~Y Ng, and Christopher Potts. 2013.
\newblock Recursive deep models for semantic compositionality over a sentiment treebank.
\newblock In \emph{Proceedings of the Conference on Empirical Methods in Natural Language Processing}.

\bibitem[{Strubell et~al.(2018)Strubell, Verga, Andor, Weiss, and McCallum}]{strubell-etal-2018-linguistically}
Emma Strubell, Patrick Verga, Daniel Andor, David Weiss, and Andrew McCallum. 2018.
\newblock Linguistically-informed self-attention for semantic role labeling.
\newblock In \emph{Proceedings of the Conference on Empirical Methods in Natural Language Processing}.

\bibitem[{Tai et~al.(2015)Tai, Socher, and Manning}]{tai2015improved}
Kai~Sheng Tai, Richard Socher, and Christopher~D Manning. 2015.
\newblock Improved semantic representations from tree-structured long short-term memory networks.
\newblock In \emph{Proceedings of the Annual Meeting of the Association for Computational Linguistics and the International Joint Conference on Natural Language Processing (Volume 1: Long Papers)}.

\bibitem[{Team et~al.(2023)Team, Anil, Borgeaud, Alayrac, Yu, Soricut, Schalkwyk, Dai, Hauth, Millican et~al.}]{team2023gemini}
Gemini Team, Rohan Anil, Sebastian Borgeaud, Jean-Baptiste Alayrac, Jiahui Yu, Radu Soricut, Johan Schalkwyk, Andrew~M Dai, Anja Hauth, Katie Millican, et~al. 2023.
\newblock Gemini: a family of highly capable multimodal models.
\newblock \emph{arXiv preprint arXiv:2312.11805}.

\bibitem[{van Schijndel et~al.(2013)van Schijndel, Exley, and Schuler}]{van2013model}
Marten van Schijndel, Andy Exley, and William Schuler. 2013.
\newblock A model of language processing as hierarchic sequential prediction.
\newblock \emph{Topics in cognitive science}, 5(3):522--540.

\bibitem[{Vaswani et~al.(2017)Vaswani, Shazeer, Parmar, Uszkoreit, Jones, Gomez, Kaiser, and Polosukhin}]{vaswani2017attention}
Ashish Vaswani, Noam Shazeer, Niki Parmar, Jakob Uszkoreit, Llion Jones, Aidan~N. Gomez, \L{}ukasz Kaiser, and Illia Polosukhin. 2017.
\newblock Attention is all you need.
\newblock In \emph{Proceedings of the International Conference on Neural Information Processing Systems}.

\bibitem[{Wang et~al.(2019)Wang, Lee, and Chen}]{wang-etal-2019-tree}
Yaushian Wang, Hung-Yi Lee, and Yun-Nung Chen. 2019.
\newblock Tree transformer: Integrating tree structures into self-attention.
\newblock In \emph{Proceedings of the Conference on Empirical Methods in Natural Language Processing and the International Joint Conference on Natural Language Processing}.

\bibitem[{Warstadt et~al.(2020)Warstadt, Parrish, Liu, Mohananey, Peng, Wang, and Bowman}]{warstadt2020blimp}
Alex Warstadt, Alicia Parrish, Haokun Liu, Anhad Mohananey, Wei Peng, Sheng-Fu Wang, and Samuel~R Bowman. 2020.
\newblock Blimp: The benchmark of linguistic minimal pairs for english.
\newblock \emph{Transactions of the Association for Computational Linguistics}, 8:377--392.

\bibitem[{Williams et~al.(2018)Williams, Nangia, and Bowman}]{williams-etal-2018-broad}
Adina Williams, Nikita Nangia, and Samuel Bowman. 2018.
\newblock A broad-coverage challenge corpus for sentence understanding through inference.
\newblock In \emph{Proceedings of the Conference of the North {A}merican Chapter of the Association for Computational Linguistics: Human Language Technologies, Volume 1 (Long Papers)}.

\bibitem[{Xia et~al.(2024)Xia, Gao, Zeng, and Chen}]{xiasheared}
Mengzhou Xia, Tianyu Gao, Zhiyuan Zeng, and Danqi Chen. 2024.
\newblock Sheared llama: Accelerating language model pre-training via structured pruning.
\newblock In \emph{Proceedings of the International Conference on Learning Representations}.

\bibitem[{Yanaka et~al.(2019)Yanaka, Mineshima, Bekki, Inui, Sekine, Abzianidze, and Bos}]{yanaka2019can}
Hitomi Yanaka, Koji Mineshima, Daisuke Bekki, Kentaro Inui, Satoshi Sekine, Lasha Abzianidze, and Johan Bos. 2019.
\newblock Can neural networks understand monotonicity reasoning?
\newblock In \emph{Proceedings of the ACL Workshop BlackboxNLP: Analyzing and Interpreting Neural Networks for NLP}.

\bibitem[{Yao et~al.(2021)Yao, Peng, Papadimitriou, and Narasimhan}]{yao-etal-2021-self}
Shunyu Yao, Binghui Peng, Christos Papadimitriou, and Karthik Narasimhan. 2021.
\newblock Self-attention networks can process bounded hierarchical languages.
\newblock In \emph{Proceedings of the Annual Meeting of the Association for Computational Linguistics and the International Joint Conference on Natural Language Processing (Volume 1: Long Papers)}.

\bibitem[{Younger(1967)}]{younger1967recognition}
Daniel~H Younger. 1967.
\newblock Recognition and parsing of context-free languages in time n3.
\newblock \emph{Information and control}, 10(2):189--208.

\end{thebibliography}
